\documentclass[sigconf]{acmart}

\AtBeginDocument{%
  }

\copyrightyear{2025}
\acmYear{2025}
\setcopyright{acmlicensed}\acmConference[KDD '25]{Proceedings of the 31st ACM SIGKDD Conference on Knowledge Discovery and Data Mining V.1}{August 3--7, 2025}{Toronto, ON, Canada}
\acmBooktitle{Proceedings of the 31st ACM SIGKDD Conference on Knowledge Discovery and Data Mining V.1 (KDD '25), August 3--7, 2025, Toronto, ON, Canada}
\acmDOI{10.1145/3690624.3709212}
\acmISBN{979-8-4007-1245-6/25/08}

\usepackage{CJKutf8}
\usepackage{balance}
\usepackage{multirow}
\usepackage{booktabs} 
\usepackage{xcolor}
\usepackage{soul}
\usepackage{tabularx} 
\usepackage{array}
\usepackage{subcaption}
\usepackage{hhline}
\usepackage{graphicx} 
\usepackage{float}
\usepackage{amsmath}
\usepackage{amsthm} 
\newtheorem{assumption}{Assumption}
\begin{document}
\begin{CJK}{UTF8}{gbsn}

\title{Boosting Explainability through Selective Rationalization in Pre-trained Language Models}

\author{Libing Yuan}
\authornote{Both authors contributed equally to this research.}
\affiliation{
        \department{School of Computer Science and Information Engineering}
        \institution{Hefei University of Technology}
	\city{Hefei}
	\country{China}}
\email{libingyuan@mail.hfut.edu.cn}

\author{Shuaibo Hu}
\authornotemark[1]
\affiliation{
        \department{School of Computer Science and Information Engineering}
        \institution{Hefei University of Technology}
	\city{Hefei}
	\country{China}}
\email{shuaibohu@mail.hfut.edu.cn}
    
\author{Kui Yu}
\authornote{Corresponding author.}
\affiliation{
    \department{School of Computer Science and Information Engineering}
	\institution{Hefei University of Technology}
	\city{Hefei}
	\country{China}}
\email{yukui@hfut.edu.cn}

\author{Le Wu}
\affiliation{
        \department{School of Computer Science and Information Engineering}
        \institution{Hefei University of Technology}
	\city{Hefei}
	\country{China}}
\email{lewu.ustc@gmail.com}


    


\begin{abstract}
The widespread application of pre-trained language models (PLMs) in natural language processing (NLP) has led to increasing concerns about their explainability. 
Selective rationalization is a self-explanatory framework that selects human-intelligible input subsets as rationales for predictions.
Recent studies have shown that applying existing rationalization frameworks to PLMs will result in severe degeneration and failure problems, producing sub-optimal or meaningless rationales.
Such failures severely damage trust in rationalization methods and constrain the application of rationalization techniques on PLMs.
In this paper, we find that the homogeneity of tokens in the sentences produced by PLMs is the primary contributor to these problems.
To address these challenges, we propose a method named Pre-trained Language Model's Rationalization (PLMR), which splits PLMs into a generator and a predictor to deal with NLP tasks while providing interpretable rationales.
The generator in PLMR also alleviates homogeneity by pruning irrelevant tokens, while the predictor uses full-text information to standardize predictions.
Experiments conducted on two widely used datasets across multiple PLMs demonstrate the effectiveness of the proposed method PLMR in addressing the challenge of applying selective rationalization to PLMs.
Codes: https://github.com/ylb777/PLMR.
\end{abstract}

\ccsdesc[500]{Computing methodologies~Natural language processing}
\ccsdesc[300]{Computing methodologies~Knowledge representation and reasoning}
\ccsdesc[500]{Computing methodologies~Machine learning approaches}
\keywords{Pre-trained Language Models; Explainability; Rationalization}


\maketitle
\section{Introduction}
\label{intro}
\begin{figure}[htbp]
  \centering
  \includegraphics[width=\linewidth]{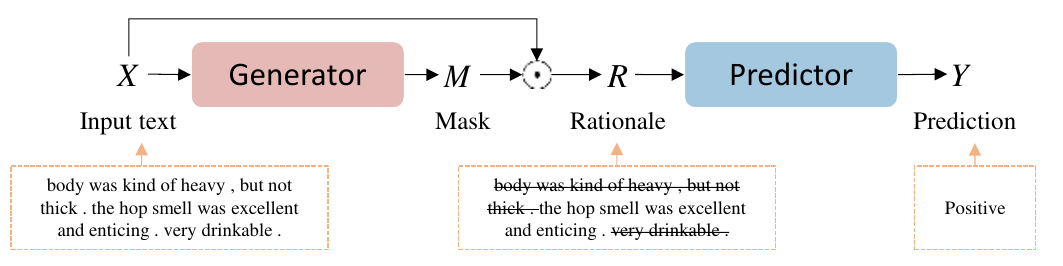}
  \caption{The selective rationalization framework RNP.}
  \label{figure-RNP}
\end{figure}

The widespread application of deep learning models, particularly pre-trained language models (PLMs), across critical fields in natural language processing (NLP) has led to increasing concerns about their explainability \cite{gurrapu2023rationalization,danilevsky-etal-2020-survey}. 
Selective rationalization, as an interpretable method, has received continuous research in the field in recent years \cite{lyu2024towards,jacovi2020towards}. 
Lei et al. \cite{lei2016rationalizing} were the first to propose this framework for rationalizing neural predictions (RNP), which consists of a generator and a predictor. 
The generator selects a human-intelligible subset of the entire input sentence as a rationale, while the predictor then makes a judgment only based on this rationale, ensuring that the explanation is faithful, as shown in Figure \ref{figure-RNP}. 

The vanilla RNP framework in Figure \ref{figure-RNP} suffers from rationalization degeneration and failure problems \cite{a2r,liu2022fr,zheng-etal-2022-irrationality}.  
The question with rationalization degeneration lies in its capacity to yield lower prediction loss even with a poor quality of rationale.
For example, the final prediction is correct even if the rationale, as shown in Figure 2, is not as relevant to the gold annotations. 
The rationalization failure \cite{jacovi2021aligning,hu2024learning} goes further, as the chosen rationale was completely absurd, but the label can still be accurately predicted.
Many methods have been proposed to alleviate these issues \cite{a2r, dare, hu2024learning}.
In these studies, researchers focused on enhancing the rationalization framework, so their experiments utilized relatively simple networks as encoders, such as GRU \cite{cho2014learning} and LSTM \cite{lstm}. 
This assumes that the framework remains valid when the encoder is altered.

Although these methods have achieved improvements, recent studies \cite{chen2022can, liu2024d} have indicated that the previous framework for selective rationalization will not work if we replace the encoder with the PLM.
For example, as shown in Table \ref{tab:bert-gru}, using a three-layer GRU can provide better token representations compared to a single-layer GRU, allowing for the selection of more accurate rationales.
Contrarily, PLMs with more extensive parameters, which were expected to learn better representations \cite{lewis2019bart,raffel2020exploring}, actually yield poor-quality rationales. 
When using the same selective rationalization framework in Table \ref{tab:bert-gru}, we can observe that employing BERT \cite{devlin2018bert} does not offer better rationales compared to using GRU; instead, the quality of rationale becomes worse.

\begin{figure}[htbp]
  \centering
  \includegraphics[width=\linewidth]{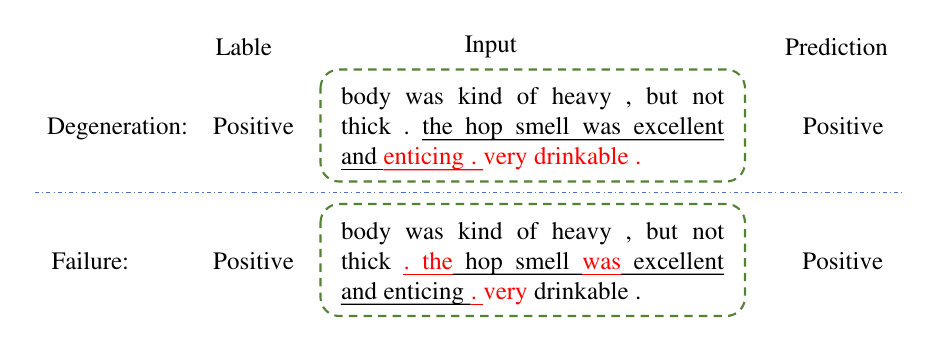}
  \caption{Two examples illustrate the rationalization degeneration and rationalization failure.  
  Human-annotated rationales are \ul{underlined}. Rationales from the rationalization framework are highlighted in \textcolor{red}{red}.
  }
  \label{figure-failure}
\end{figure}

\begin{table}[htbp]
\caption{The F1 scores of selection rationale using different methods of BERT or GRU on Aroma aspects of the BeerAdvocate dataset. The experimental settings are provided in appendix \ref{table1-detail}.}
\label{tab:bert-gru}
\begin{tabular}{c|ccccc}
\hline
Method       & RNP  & INVRAT & FR   & CR   & G-RAT  \\ \hline
GRU(1-layer) & 41.8 & 44.1   & 50.2 & -    & 52.9 \\ \hline
GRU(3-layers) & 44.9 & 45.0   & 53.7 & -    & 56.5 \\ \hline
BERT         & 32.0   & 33.2   & 39.5 & 49.1 & 38.4 \\ \hline
\end{tabular}
\end{table}

Thus, the questions naturally arise: what factors lead to the poor performance of the selective rationalization framework within PLMs, and what strategies can improve it? 
Motivated by these two core concerns, we conducted a series of analyses and proposed a method called \textbf{P}re-trained \textbf{L}anguage \textbf{M}odel's \textbf{R}ationalization (\textbf{PLMR}) that can work on PLMs.  
Our contributions are as follows:

First, we conduct experiments to validate the severe rationalization degeneration and failure problems that occur when we apply the BERT model (a commonly used PLM) to the current rationalization frameworks. 
We then deeply analyze why these two problems occur when using the BERT model. 
We find that the over-learning of contextual information of PLMs causes the input tokens to have homogeneous representations, leading to severe rationalization degeneration and failure problems.

Second, based on the analysis above, we propose a novel approach named PLMR.
PLMR Truncates the PLM into two independent parts: the rationale generation part and the prediction part. 
But from a perspective outside of PLM, the two parts still exist as a whole.
In order to further reduce the learning of context-free information, PLMR cut out irrelevant context before rationale generation. 
In the prediction part, PLMR utilizes full-text information to regularize the predictions of the predictor, further improving the performance of the model.

Thirdly, the F1 score for rationale selection using PLMR is up to 9\% higher than methods using GRU and up to 17\% higher than previous methods also using PLMs.
These results indicate that our PLMR can use PLM in rationalization and effectively address issues of rationalization degeneration and failure, making an essential step towards rationalizing the prediction of PLMs for their explainability.

\section{RELATED WORK}
The base selective rationalization framework named RNP \cite{lei2016rationalizing} uses only the rationale extracted by the generator as justifications to make predictions, thereby ensuring the interpretation of the cooperative model is faithful \cite{lyu2024towards}.
Such cooperative frameworks between generator and predictor are hard to optimize as well as train. 
To address this problem, Bao et al. \cite{bao-etal-2018-deriving} used Gumbel-softmax to re-parameterize gradient estimates. Bastings et al. \cite{bastings-etal-2019-interpretable} employed a rectified Kumaraswamy distribution to replace Bernoulli sampling.

Then it was found that rationalization degeneration and failure problems occur due to the predictor using only the rationale for prediction. To address the degeneration problem, a series of studies used additional information to regularize the predictor.
3PLAYER \cite{3player} adds a complementary predictor that uses text not selected as the rationale.
A2R \cite{a2r} uses soft attention from the generator to input full-text information into the predictor.
DMR \cite{dmr} aligns the feature and output distributions of the rationales with the full-text input.
DARE \cite{dare} improves rationale representations by reducing the mutual information between rationale and non-rationale parts of the input.
FR \cite{liu2022fr} uses the same encoder between the generator and the predictor to convey information.
DAR \cite{liu2024enhancing} utilizes an auxiliary module pretrained on the full input to align the selected rationale and the original input discriminatively.
Meanwhile, some work addresses the degeneration problem from a causal perspective.
INVRAT \cite{invrat} uses a game-theoretic approach to constrain the output of rationale across multiple environments.
Inter-RAT \cite{inter-rat} uses a backdoor adjustment method theory to remove spurious correlations.
MCD \cite{liu2024d} proposes the Minimum Conditional Dependence (MCD) criterion to uncover causal rationales.
MRD
Rationalization failure undermines the user's trust more severely by selecting meaningless rationale.
G-RAT \cite{hu2024learning} addresses rationalization failure problems by using a guidance module to regularize the selection of the generator.

The above approach typically uses simple recurrent models such as GRU \cite{a2r,liu2022fr,hu2024learning,inter-rat,liu2024mmi} and LSTM \cite{3player}. 
Recent experiments \cite{chen2022can,liu2022fr} have shown that these rationalization frameworks with PLMs lead to the poor quality of rationales.
Although some studies have been conducted using PLMs \cite{cr,yuetowards}, the quality of rationales remains inferior compared to methods utilizing GRU.
Our work analyzes the reasons for the problems of degeneration and failure of selective rationalization with PLMs and proposes a framework suitable for PLMs.

\section{PRELIMINARIES}
\label{preliminaries}
\textbf{Selective Rationalization.}
We consider a text classification task, where the input text is $X=\left[x_{1}, x_{2}, \cdots, x_{n}\right]$, with $x_{i}$ representing the $i$-th token and $n$ representing the number of tokens in the text. $Y$ is the label corresponding to $X$. The selective rationalization framework consists of a generator $g(\cdot)$ and a predictor $p(\cdot)$, with $\theta_{g}$ and $\theta_{p}$ representing the parameters of the generator and the predictor, respectively. In the training set $(X, Y) \in D$, the rationale is unknown. The goal of the generator is to learn a sequence of binary mask $M=\left[m_{1}, \cdots, m_{n}\right] \in\{0,1\}^{n}$ from the input $X$, and then use $M$ to generate a subset of the input text as the rationale $R$:
\begin{equation}
  R=M \odot X=\left[m_{1} x_{1}, \cdots, m_{n} x_{n}\right].
\end{equation}
Subsequently, the predictor uses $R$ to perform the text classification task while computing the task loss $L_{\text {task }}$ for the entire select-then-predict model. Finally, the collaborative optimization process of the generator and predictor is as follows:
\begin{equation}
  \min _{\theta_{g}, \theta_{p}} E_{\substack{X, Y \sim D \\ M \sim g\left(X\right)}}\left[L_{\text {task }}\left(p\left(M \odot X\right), Y\right)\right].
  \label{eq:loss_task}
\end{equation}
To ensure that the rationale selected by the generator is understandable to humans, we aim to choose short and coherent subsets as the rationale. To achieve this goal, we adopt the constraint methods used in most previous research:
\begin{equation}
  L_{s}=\lambda_{1}\left|\alpha-\frac{1}{n}\sum_{i=0}^{n}M_{i}\right|+\lambda_{2} \sum_{i=1}^{n}\left|M_{i}-M_{i-1}\right|.
  \label{eq:ls}
\end{equation}
The first term uses a predefined sparsity $\alpha \in[0,1]$ to control the proportion of the rationale selected, while the second term ensures that the rationale is as coherent as possible. Therefore, Equation \ref{eq:loss_task} can be rewritten as:
\begin{equation}
  \min _{\theta_{g}, \theta_{p}} E_{\substack{X, Y \sim D \\ M \sim g\left(X\right)}}\left[L_{\text {task }}\left(p\left(M \odot X\right), Y\right)+L_{s}\right].
  \label{eq:loss_task2}
\end{equation}

\section{Analyzing Degeneration and Failure in Rationalization with PLMs}
In Section \ref{intro}, both previous studies and our experiments indicate that using PLMs in a rationalization framework can result in severe rationalization degeneration
and failure problems, providing unsatisfactory explanations. This section will first validate the presence of more severe rationalization degeneration and failure phenomena. Subsequently, we will identify the root causes of these issues by analyzing the operational mechanism of PLMs. 
Our primary experiments are performed using BERT-base-uncased, which has 12 layers (transformer \cite{vaswani2017attention} blocks), 12 attention heads, and 110 million parameters.

\subsection{Rationalization Degeneration and Failure}
\subsubsection{Rationalization Degeneration.}
For two subsets, $X_{1}$ and $X_{2}$, within the input text $X$, where $X_{1}$ is the golden rationale corresponding to label $Y$, the high correlation between $X_{1}$ and $X_{2}$ results in a spurious correlation between $X_{2}$ and $Y$. This spurious correlation is the possible cause of degeneration. The strength of the correlation between $X_{1}$ and $X_{2}$ reflects the extent of the spurious correlation. Therefore, we demonstrate through the following experiments that using pre-trained language models (PLMs) in a rationalization framework leads to more severe degeneration.
\begin{figure}[htbp]
    \centering
    \begin{subfigure}{0.15\textwidth}
        \centering
        \includegraphics[width=\textwidth]{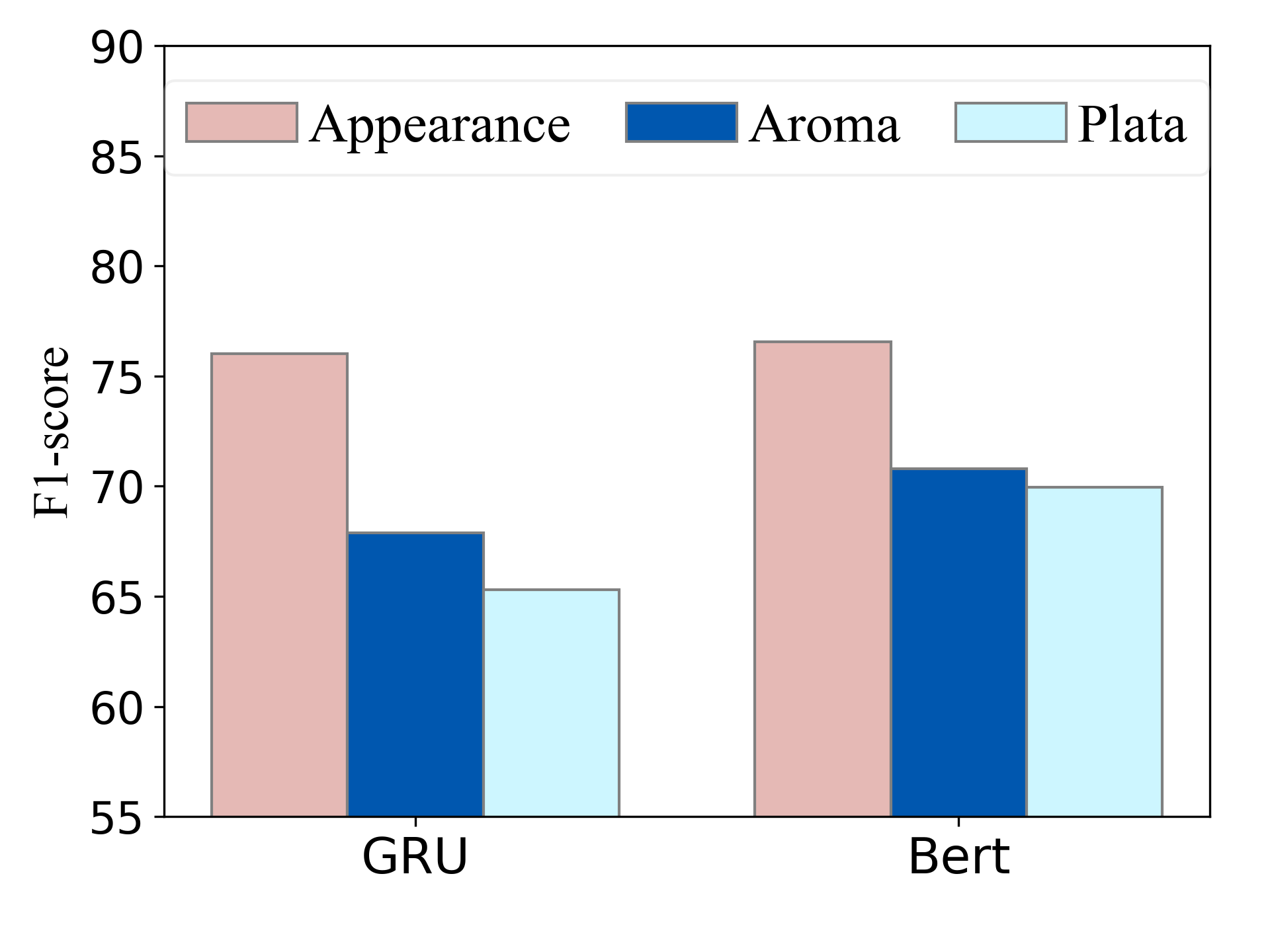}
        \caption{appearance}
        \label{aspect0}
    \end{subfigure}
    \hfill
    \begin{subfigure}{0.15\textwidth}
        \centering
        \includegraphics[width=\textwidth]{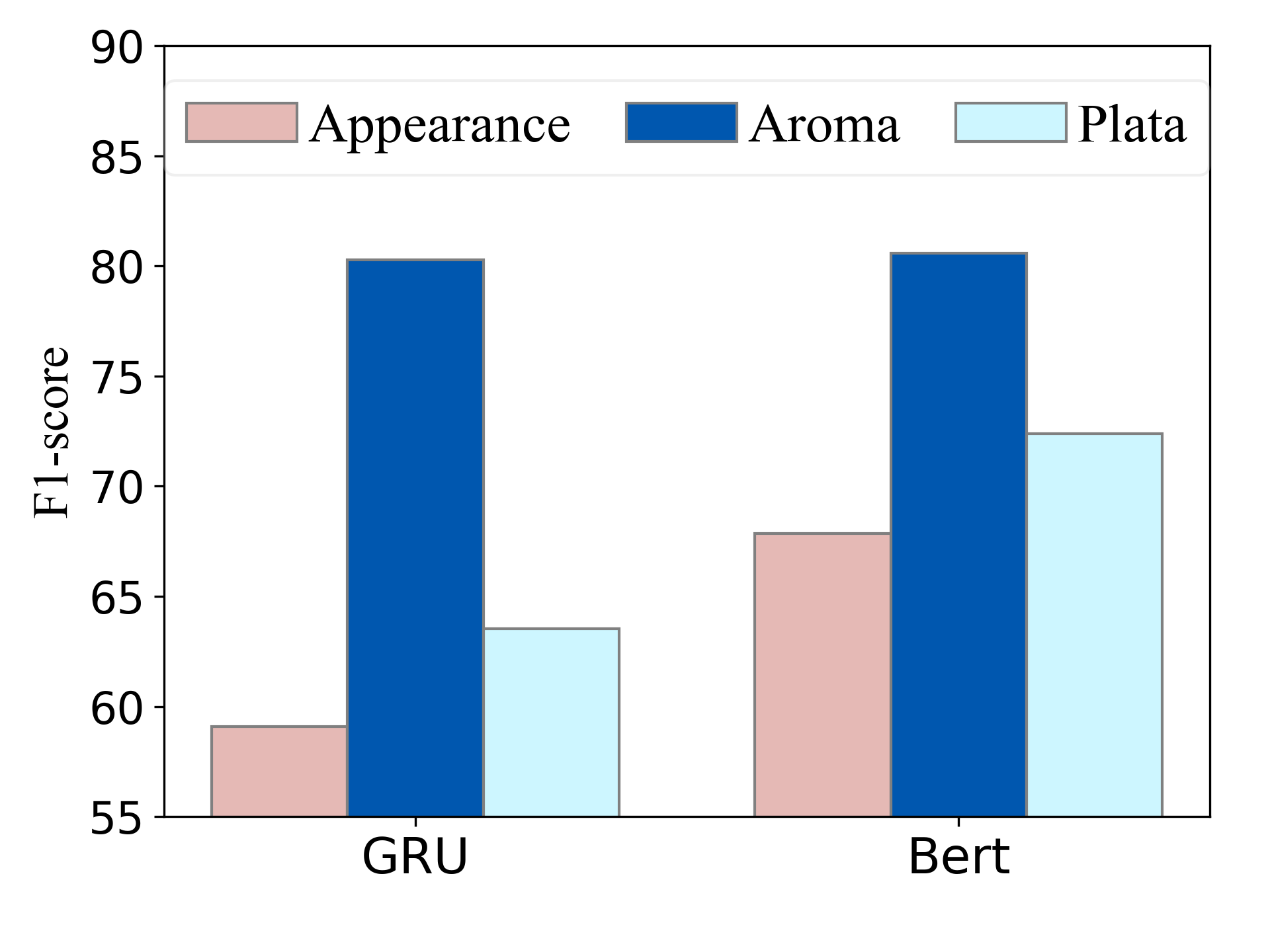}
        \caption{aroma}
        \label{aspect1}
    \end{subfigure}
    \hfill
    \begin{subfigure}{0.15\textwidth}
        \centering
        \includegraphics[width=\textwidth]{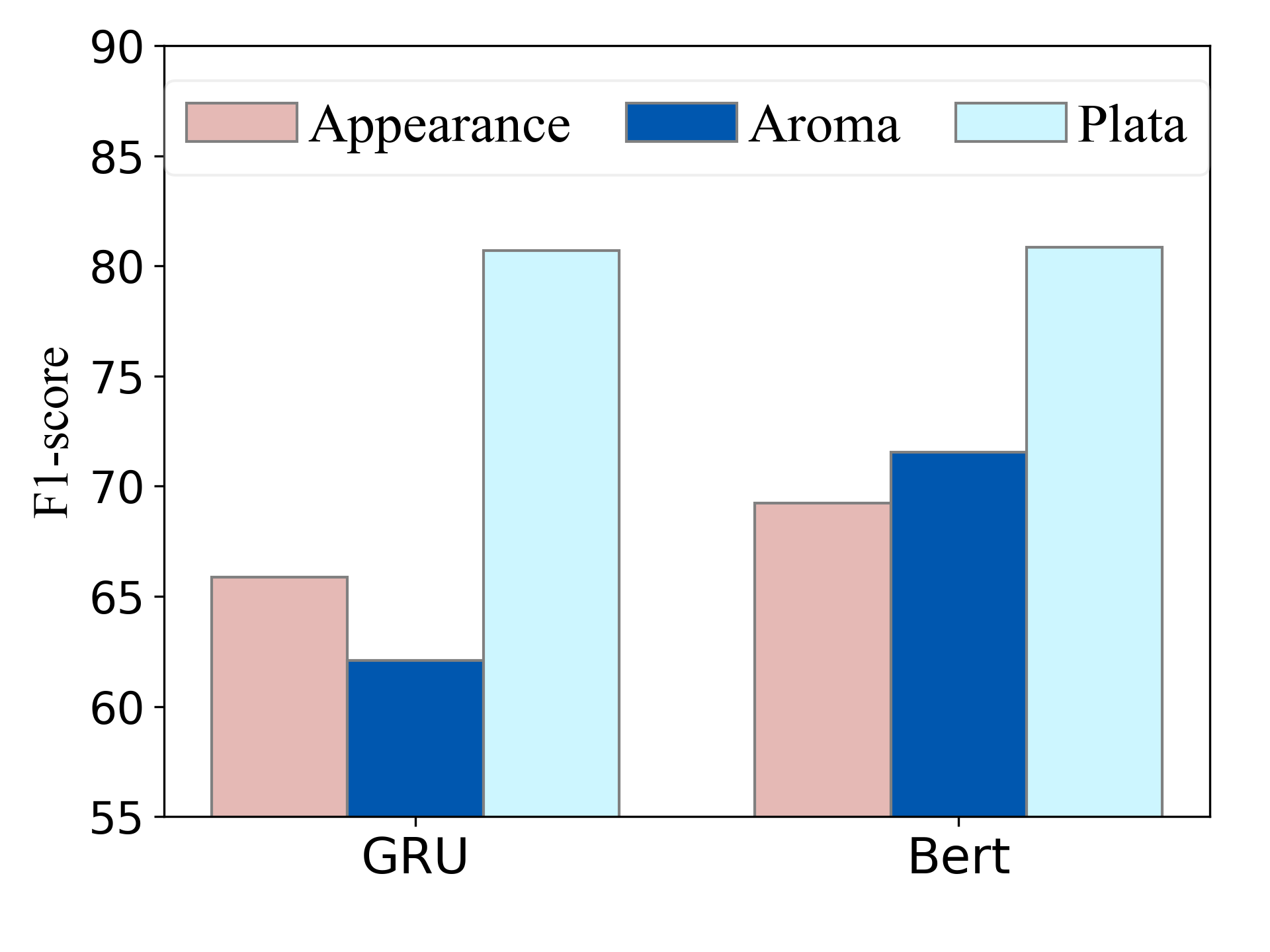}
        \caption{palate}
        \label{aspect2}
    \end{subfigure}
    \caption{Comparison of spurious correlation strength in BERT and GRU representations. 
    }
    \label{figure-Bert-GRU}
\end{figure}

In the BeerAdvocate dataset, each text contains multiple rationales (appearance, aroma, palate) and corresponding labels. First, we train a classifier using a GRU on the appearance aspect and its corresponding labels. Then we use this classifier to predict the aroma and palate aspects. 
As shown in Figure \ref{aspect0}, the GRU classifier can effectively predict the untrained aroma and palate aspects, which demonstrates a significant correlation between different rationales. These correlations result in spurious associations between aroma, palate, and the label of appearance, ultimately causing the rationalization model to select spurious rationales.
Next, we conduct the same experiment using a BERT classifier, ensuring that the BERT and GRU classifiers had the same prediction capability (F1 score) on appearance. Comparing the results of the experiments with GRU and BERT in Figure \ref{aspect0}, we find that the BERT classifier achieved higher F1 scores on the aroma and palate aspects. The results consistently showed that, compared to GRU, BERT significantly increases the influence of spurious correlations. 
We obtain the same experimental results for the other two aspects in Figure \ref{aspect1} and \ref{aspect2}.
The increasing impact of spurious correlations leads to more severe degeneration in rationalization for PLMs like BERT, and we also provide an additional example in Appendix \ref{examples1} for a further illustration.

\subsubsection{Rationalization Failure.}
When the generator selects meaningless tokens as its rationale, the predictor should theoretically fail to make accurate predictions. However, there are occasions when the predictor still succeeds in making correct predictions.
This occurrence causes the generator to mistakenly believe these tokens support its predictions, leading to what is known as rationalization failure.
Based on the reasons for the rationalization failure mentioned above, we make the assumption to illustrate whether rationale failure occurs:
\begin{assumption}
    Given a rationalization framework, which consists of generator $g\left(\cdot \right)$ and predictor $p\left(\cdot \right)$.
For $(X, R, Y) \in D_{\text {test }}$, $g(X)=\hat{R}$, $p(\hat{R})=\hat{Y}$. Rationalization failure occurs when the learned rationale $\hat{R}$ meets the following conditions:

(1) $p(\hat{R})=\hat{Y}=Y$; The predictor $p\left(\cdot \right)$ predicts correctly using $\hat{R}$.

(2) $P=|R \cap \hat{R}| /|R| \leq \theta_{1}$; The proportion $P$ of correct tokens in the $\hat{R}$ is less than $\theta_{1}$.

(3) $\left|\hat{R}_{\text {punct, prep, pron, art, conj }}\right| /|\hat{R}| \geq \theta_{2}$; The proportion of meaningless tokens such as punctuation, prepositions, pronouns, articles, and conjunctions in $\hat{R}$ is greater than $\theta_{2}$.
\end{assumption}

Therefore, we explored the likelihood of rationalization failures in PLMs and RNNs through the selection rationalization framework.
Our experiments utilized BERT and GRU within the RNP framework. By adjusting parameters $\theta_{1}$ and $\theta_{2}$, we observed the occurrence of rationalization failures in the test set $D_{\text {test}}$. As shown in Table \ref{table-failure}, BERT-RNP experienced a significantly higher rate of rationalization failures compared to GRU-RNP. 
The result shows that using PLMs in a rationalization framework often leads to the selection of numerous meaningless tokens as rationales, resulting in a significant rationalization failure.
The experimental details are in Appendix \ref{table2-detail}.
\begin{table}[htbp]
\caption{Percentage of rationalization failure texts in the test set. Experiments on the aroma aspect of the BeerAdvocate.}
\label{table-failure}
\begin{tabular}{c|ccc|ccc}
\hline
$\theta_{1}$ & \multicolumn{3}{c|}{0.2} & \multicolumn{3}{c}{0.3} \\ \hline
$\theta_{2}$  & 0.3    & 0.4    & 0.5    & 0.3    & 0.4    & 0.5   \\ \hline
GRU-RNP    & 1.93   & 1.05   & 0.36   & 3.92   & 1.75   & 0.54  \\
BERT-RNP   & 47.71  & 30.74  & 12.40  & 69.75  & 44.93  & 17.48 \\ \hline
\end{tabular}
\end{table}

\subsection{Homogeneity Among Tokens and Clauses}
Unlike recurrent neural networks, the self-attention mechanism of the Transformer model allows each token to establish direct connections with other tokens in the sentence. This mechanism enables each token to attend to any position in the sentence, thereby capturing long-distance dependencies. However, in PLMs with multi-layered Transformer structures, each token learns information from other tokens at every layer of the Transformer. This over-learning of contextual information leads to highly similar final token representations, resulting in a lack of heterogeneity.
We refer to the phenomenon where different tokens within a sentence exhibit similar semantic information as token homogeneity. In this section, we empirically demonstrate the occurrence of homogeneity in PLMs and explain how it leads to more severe rationalization degeneration and rationalization failure problems.

To illustrate the homogeneity of token representations generated by BERT, we utilize the traditional likelihood-based variance-covariance matrix homogeneity test method \cite{box1949general} to observe the degree of discrepancy between token representations.

Given the hidden states generated by BERT, denoted as $H$, where $H$ is an $n \times p$ matrix, $n$ is the number of tokens in the sentence, and $p$ is the dimensionality of the token representations. The calculation of the variance-covariance matrix is as follows:
\begin{equation}
  H_{ij}^{\prime}=H_{ij}-\frac{1}{n} \sum_{k=1}^{n} H_{kj},
\end{equation}
\begin{equation}
  \Sigma=\frac{1}{n-1}\left(H^{\prime}\right)^{T} H^{\prime},
  \label{sigma}
\end{equation}
where $H^{\prime}$ is the centralized data matrix, $\Sigma$ in Eq.(\ref{sigma}) is the variance-covariance matrix. $\Sigma_{i i}$ represents the variance of the $i$-th dimension. Therefore, the trace of the variance-covariance matrix can intuitively reflect the degree of discrepancy among different token representations. So, a smaller trace value indicates more severe homogeneity.
\begin{equation}
  \operatorname{tr}(\Sigma)=\sum_{i=1}^{p} \Sigma_{i i}.
\end{equation}

\begin{figure}[htbp]
    \centering
    \begin{subfigure}{0.23\textwidth}
        \centering
        \includegraphics[width=\textwidth]{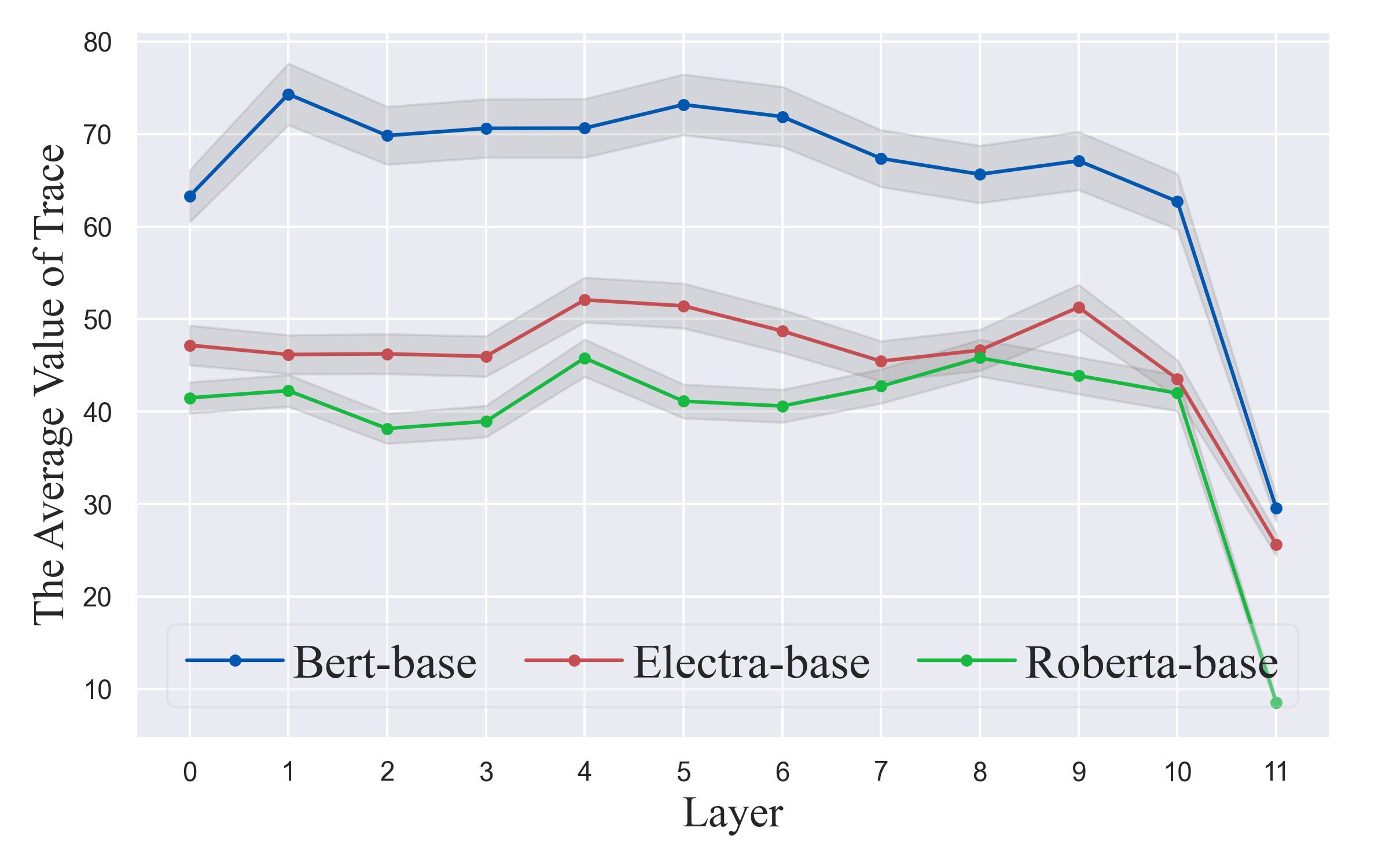}
        \caption{base-model (12-layer)}
        \label{base-model}
    \end{subfigure}
    \hfill
    \begin{subfigure}{0.23\textwidth}
        \centering
        \includegraphics[width=\textwidth]{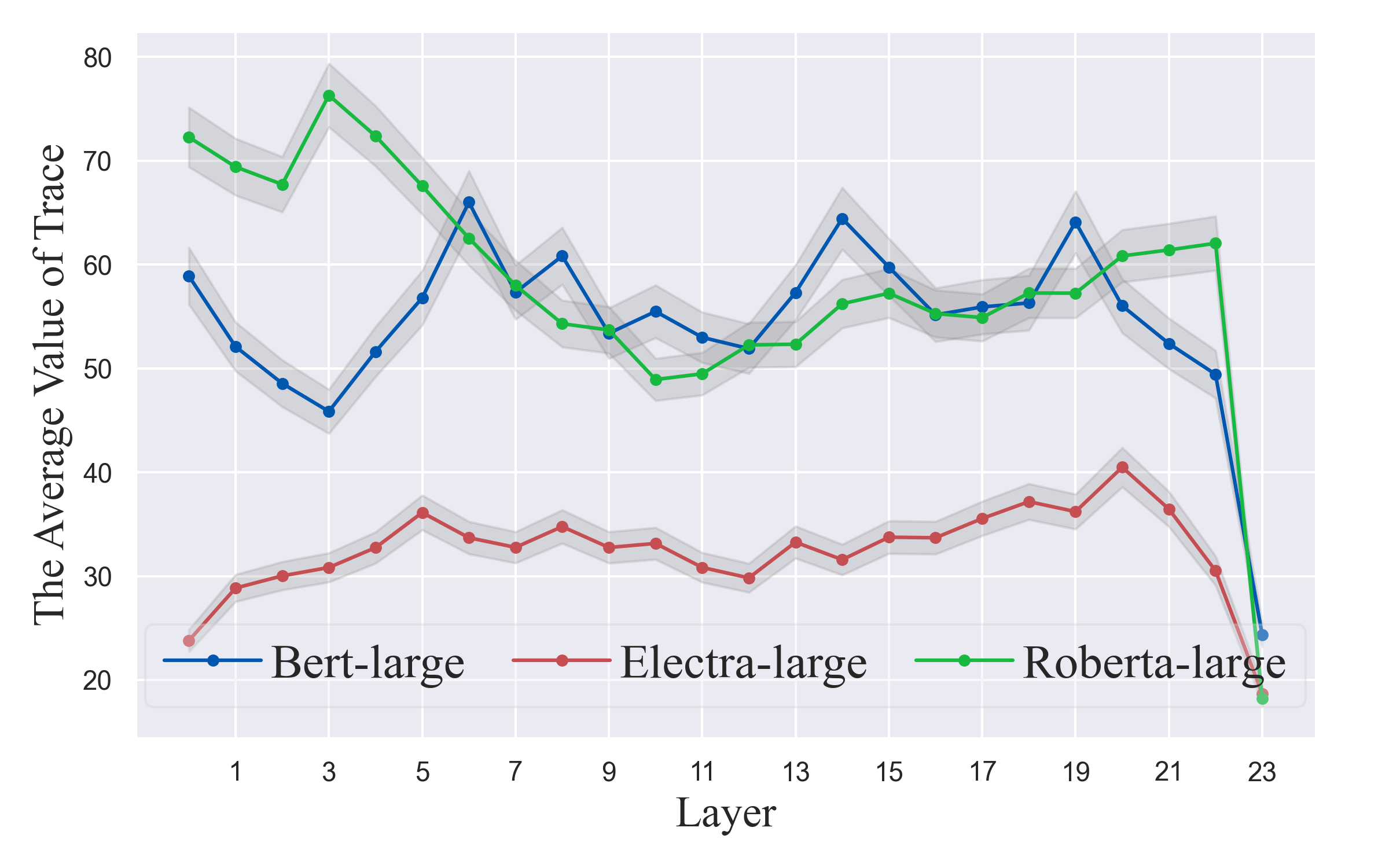}
        \caption{large-model (24-layer)}
        \label{large-model}
    \end{subfigure}
    \caption{ 
    The average trace $\operatorname{tr}(\Sigma)$ of all sentences in different layers. 
    Testing on the HotelReview dataset.
    The results are verified on three models: Bert, Electra, and Roberta.
    }
    \label{figure-trace}
\end{figure}

We conduct experiments on three types of PLMs, each including both base and large versions. Figure \ref{figure-trace} shows the trace $\operatorname{tr}(\Sigma)$ of hidden states across different transformer layers in PLMs.
The results show that the trace $\operatorname{tr}(\Sigma)$ of the final output hidden states of the PLMs are significantly smaller than those of other layers. This indicates severe homogeneity of token representations in sentences generated by PLMs.
This homogeneity implies that each token has a similar attention-weight vector in the attention heads, which means all tokens have similar attention dependencies. Appendix \ref{homogeneity} shows the distribution of attention weight vectors at different layers in Bert to support the above experiments.

Within the framework of rationale selection, the mask selection $m(x)$ needs to select an interpretable subset of tokens as rationales from the different tokens in a sentence. 
Similar to clustering tasks, we need to select the golden rationale within a sentence. 
However, the homogeneity of tokens makes this process exceptionally challenging.
On the one hand, this will lead to further correlation between different subclauses of the input text, making it easier for the generator to select sub-optimal subsets. On the other hand, token homogeneity means that the heterogeneity between different tokens is reduced, and it is challenging to distinguish rationales from other meaningless tokens, causing the generator to select meaningless tokens in the early stages of training easily. When both situations occur, powerful predictors will overfit the wrong results and make accurate predictions, leading to more severe rationalization degeneration and rationalization failure problems. 

Therefore, we understand that the more severe rationalization degeneration and failure occur due to the generator and the predictor issues. The key reason lies in the homogeneity of tokens produced by the generator. Another reason is that the predictor makes correct predictions based on incorrect rationales.
\begin{figure*}[ht]
  \centering
  \includegraphics[width=1.8\columnwidth]{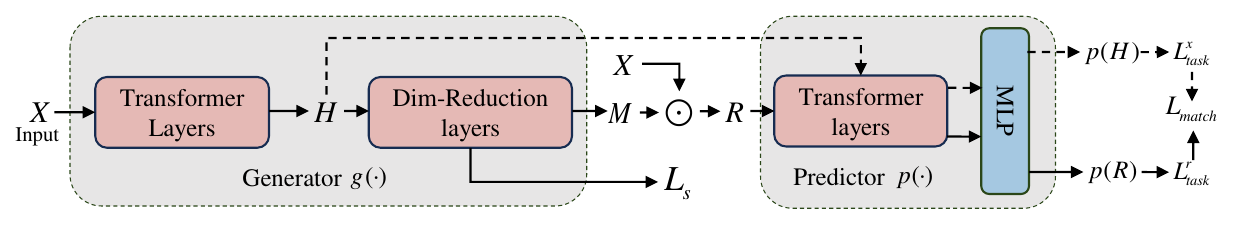}
  \caption{The proposed rationalization architecture PLMR. The dashed lines are used only during the training phase.
  }
  \label{figure-br}
\end{figure*}

\section{Methodology}
Based on the above analysis, we address these issues from the perspectives of both the generator and the predictor and propose the method PLMR.
In this section, we first describe the overall architecture of PLMR, which consists of the rationale selection and prediction modules (Section \ref{sec:overall_architecture}). 
Then we present the details of these two modules (Section \ref{sec: generator} and Section \ref{sec: predictor}).

\subsection{Overall Architecture}
\label{sec:overall_architecture}
\begin{figure}[htbp]
  \centering
  \includegraphics[width=0.9\columnwidth]{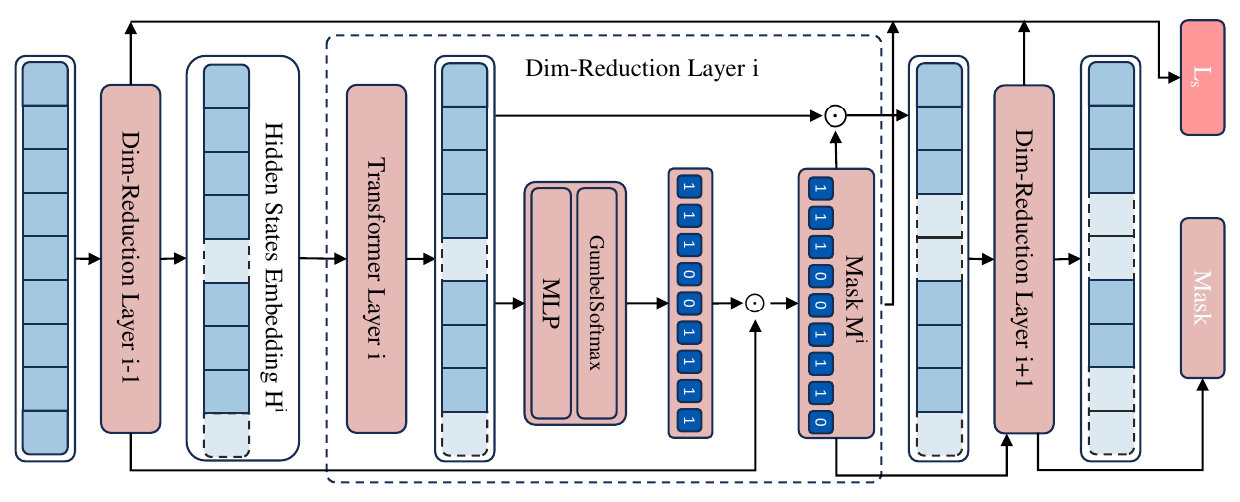}
  \caption{Detailed Design of the Dimension-Reduction Layer: The input is the hidden states $H$. The outputs are the final rationale mask $M$ and the sum of all sparsity and continuity control losses $L_{s}$ from the Dim-reduction layers.}
  \label{figure-dim}
\end{figure}

The overall architecture of the proposed method \textbf{PLMR} is illustrated in Figure \ref{figure-br}.
We divide PLMs into earlier layers of PLMs as the generator and the later layers of PLMs as the predictor. 
The generator is comprised of transformer layers and dimension reduction layers (Dim-Reduction layers).
The detailed design of the Dim-Reduction layers is shown in Figure \ref{figure-dim}.

As shown in Figure \ref{figure-br}, the generator in PLMR first inputs the text $X$ into a multi-layer transformer to learn the hidden states $H$. 
Then the hidden states $H$ are passed through the Dim-Reduction layers to generate the rationale mask $M$, from which we obtain the rationale $R=X \odot M$. 
The predictor then uses only the rationale $R$ for task prediction to ensure the explanation is faithful in the inference phase.

\subsection{Rationale Selection Module}
\label{sec: generator}
In this section, we explain the idea of the generator and describe its rationale selection process.
Traditional rationalization methods fail on PLMs because of the homogeneity among tokens. Therefore, we address this issue with the following two methods.

\subsubsection{Selecting for Token Heterogeneity}
\label{selct-generator}
In contrast to homogeneity, heterogeneity is crucial.
The generator's task is to select the golden rationale that best supports the prediction from the entire input text. 
The homogeneity of tokens within the text makes it challenging for the rationale mask selector to identify the correct tokens as the rationale accurately.
Figure \ref{figure-trace} shows that only the final representations of PLMs approach homogeneity, while the token representations generated by the intermediate transformer layers maintain better heterogeneity. 
 
Therefore, the earlier layers of PLMs are used as the generator $g\left(\cdot \right)$ to select rationales, which ensures that there is good heterogeneity between token representations when the mask selector $m(\cdot)$ determines whether tokens are rationales.
The mask $M$ of the rationale $R$ can be calculated by the following equation:
\begin{equation}
  M=g(X)=MLP\left(\text {Transformer}_{0-l}(X)\right),
\end{equation}
where $0-l$ indicates that the generator uses the first $l$ layers of transformers in the PLMs.
The token representations generated by the earlier layers in Figure \ref{figure-trace} exhibit good heterogeneity, so how many layers should we choose as generators?
(1) The generator requires sufficient transformer layers to learn better representations, enabling it to extract the correct rationale from the sentence.
(2) The predictor needs sufficient transformer layers to use the rationale for the prediction task.
Poor performance in either aspect will lead to a decline in overall performance. 
In order to balance the performance of the generator and the predictor, the number of layers of the two should be similar, so ideally, the generator should choose the first half of the transformer layer of PLMs. The specific choice will be verified in the experiment.

\subsubsection{Pruning Sequence in Dim-Reduction layers}
The methods mentioned in section \ref{selct-generator} have mitigated the impact of homogeneity on rationale selection.
However, generators' inevitable multi-layer transformer leads to different tokens acquiring excessive contextual information.
Especially for larger PLMs such as Bert-large-uncased (24-layer transformers), the generator will still contain numerous layers of transformers, which will lead to excessive fusion of information between irrelevant tokens and rationale tokens, reducing the heterogeneity between tokens.
To tackle this problem, we propose the Dim-Reduction layers to prune irrelevant tokens. 
Since the Transformer model processes all input sequence tokens in parallel, context pruning can be seen as reducing the sequence dimension.

As shown in Figure \ref{figure-br}, the Transformer layers in the generator are divided into two parts.
One part consists of standard Transformer layers that generate the hidden states $H_{k}$. 
\begin{equation}
  H_{k}=\text { Transformer }_{0-k}(X).
\end{equation}
The other part is the Dim-Reduction layers. 
In each Dim-Reduction layer, a proportion of tokens are pruned.
Figure \ref{figure-dim} illustrates the detailed design of the Dim-Reduction layers. For the $i$-th Dim-Reduction layer, the transformer layer learns the hidden state $H_{i-1}$ of the previous layer to get $H_{i}^{\prime}$. 
\begin{equation}
  H_{i}^{\prime}=\operatorname{Transformer}\left(H_{i-1}\right).
\end{equation}
Then to predict which tokens need to be pruned, we employ the MLP as a mask predictor after the transformer layer. 
This mask predictor utilizes the hidden states $H_{i}$ of the input sequence to predict whether the corresponding tokens will be pruned or retained, with an output of 0 indicating pruning and 1 indicating retaining.
The mask predictor's process of pruning hidden states 
$H_{i}$ is a binary prediction, which is non-differentiable and thus cannot be optimized through backpropagation.
Following prior work, we employ the reparameterization trick using Gumbel-Softmax for sampling.
Therefore, we can compute the mask $M_{i}^{\prime}=\operatorname{softmax}\left(M L P\left(H_{i}^{\prime}\right)\right)$ for pruning in this Dim-Reduction layer.
By optimizing this mask predictor, irrelevant tokens that do not significantly contribute to label prediction will be pruned.

The value 0 in the mask $M_{i}^{\prime}$ represents the corresponding tokens that the $i$-th Dim-Reduction layer should prune.
The tokens pruned by the previous Dim-Reduction layer should also be pruned by the next layer. 
Then we get the final pruning mask $M_{i}=M_{i}^{\prime} \odot M_{i-1}$.
The computation of the mask $M_{i}$ can be written as:
\begin{equation}
  \begin{array}{l}
M_{i}=\operatorname{softmax}\left(M L P\left(H_{i}^{\prime}\right)\right) \odot M_{i-1} \\
\text { where } M L P=\operatorname{Linear}(L N)
\end{array},
\end{equation}
where LN refers to the layer normalization.
Finally, we multiply the mask $M_{i}$ with $H_{i}^{\prime}$ to get the hidden state $H_{i}$ of the Dim-Reduction layer. 
\begin{equation}
  H_{i}=H_{i}^{\prime} \odot M_{i}.
\end{equation}
If this layer is the last layer of the Dim-Reduction layers, $M_{i}$ is the final rationale mask $M$.
Existing methods used $L_{s}$ as shown in Eq.(\ref{eq:ls}) in Section \ref{preliminaries} to control the brevity and continuity of the rationale. 
PLMR consists of multiple Dim-Reduction layers, each of which needs to control the proportion of tokens pruned.
Similar to previous research, we control the sparsity $\alpha_{j}$ of tokens retained in the $j$-th Dim-Reduction layer. The value of $\alpha$ should decrease progressively across multiple layers, ensuring that each layer prunes a certain number of less relevant tokens. The final layer's $\alpha$ corresponds to the sparsity of rationale tokens that need to be selected. In this paper, we set the variation of $\alpha$ to be linear change. 
Additionally, we apply a continuity control at each Dim-Reduction layer. 
Therefore, we compute the sparsity constraint and the continuity constraint for each of the Dim-Reduction layers and compute the mean value as the constraint term $L_{s}$:
\begin{equation}
  L_{s}=\frac{1}{m} \sum_{j=0}^{m}\left(\lambda_{1}\left|\alpha_{j}-\frac{1}{n} \sum_{i=0}^{n} M_{i}^{j}\right|+\lambda_{2} \sum_{i=0}^{n}\left|M_{i}^{j}-M_{i-1}^{j}\right|\right),
\end{equation}
where $m$ represents the number of Dim-Reduction layers, 
$\alpha_{j}$ denotes the proportion of tokens retained after pruning in the $j$-th Dim-Reduction layer, and $M_{i}^{j}$ indicates the mask for the $i$-th token in the $j$-th Dim-Reduction layer.

\subsection{Rationale Prediction Module}
\label{sec: predictor}
In this section, we discuss how PLMR prevents the predictor from overfitting the erroneous rationales produced by the generator, thereby enhancing the quality of the explanations.

\subsubsection{Regularizing Predictions} 
In Figure \ref{figure-br}, the predictor is composed of residual transformer layers and an MLP. The transformer layers learn the representation of the rationale, and the final output is obtained by applying average pooling to the sentence representation. The MLP then uses this output for prediction. 
Formally, we denote $L(p(X); Y)$ as the cross-entropy loss.
First, we use the rationale for prediction to get the prediction loss $L_{\text {task }}^{r}$: 
\begin{equation}
  L_{\text {task }}^{r}=L(p(M \odot X) ; Y),
\end{equation}
where $M$ is the rationale mask generated by the Dim-Reduction layers. Meanwhile, the predictor uses the full-text hidden state $H$ for prediction and calculates the full-text prediction loss as $L_{\text {task }}^{x}$.
\begin{equation}
  L_{\text {task }}^{x}=L(p(H)) ; Y).
\end{equation}
Finally, we use $L_{\text {match }}$ as a regularizer for generating the rationale:
\begin{equation}
  L_{\text {match }}=\lambda h\left(L_{\text {task }}^{r}-L_{\text {task }}^{x}\right),
  \label{l_match}
\end{equation}
where the function $h(x)$ is monotonically increasing and $\lambda>0$. 

\subsubsection{The Objective of $L_{\text {task }}^{x}$ and $L_{\text {match }}$}
\label{Objective of L}
Many studies \cite{a2r,liu-etal-2023-mgr,liu2024d} use additional information to guide the predictor. Inspired by this, we use $L_{\text {task }}^{x}$ to ensure the predictor utilizes full-text representations for prediction.
Early in training, when the quality of learned rationales is poor, we want $L_{\text {task }}^{r}$ and $L_{\text {task }}^{x}$ to be similar. As training progresses and rationale quality improves, we want the rationales to make more accurate predictions, $L_{\text {task }}^{r} \leq L_{\text {task }}^{x}$. We achieve this by setting the function $h(x)$ in the regularization term $L_{\text {match }}$ to be monotonically increasing. The detailed theoretical analysis is in Appendix \ref{proof-lmatch}.

\subsection{Optimization}
By combining all the above equations, the total objective of our rationale module is as follows:
\begin{equation}
  \min _{\theta_{g}, \theta_{p}} E_{\substack{X, Y \sim D \\ M \sim g\left(X\right)}}\left[L_{\text {task }}^{r}+L_{\text {task }}^{x}+L_{\text {match }}+L_{s}\right].
\end{equation} 
During the training phase, we use the Adam optimizer \cite{KingBa15} to optimize the above objective.
Empirical evaluation showed that these losses were of comparable magnitudes.
Therefore, simple averaging was chosen to maintain simplicity and robust performance.
During the inference phase, the predictor only uses the rationale generated by the generator for prediction.

\begin{table*}[htbp]
\caption{Comparison with Bert-based Methods on BeerAdvocate Dataset. “ * ” represents the results from the paper of CR \cite{cr}.}
\label{tab:bert-result}
\begin{tabularx}{0.95\textwidth}{c|*{5}{>{\centering\arraybackslash}X}|*{5}{>{\centering\arraybackslash}X}|*{5}{>{\centering\arraybackslash}X}}
\hline
\multirow{2}{*}{Methods} & \multicolumn{5}{c|}{Appearance}                                                                     & \multicolumn{5}{c|}{Aroma}                                                                                 & \multicolumn{5}{c}{Palate}                                                                                   \\ \cline{2-16} 
                         & S  & \multicolumn{1}{c|}{ACC}  & P             & \multicolumn{1}{c|}{R}             & F1            & S  & \multicolumn{1}{c|}{ACC}           & P           & \multicolumn{1}{c|}{R}             & F1            & S  & \multicolumn{1}{c|}{ACC}           & P             & \multicolumn{1}{c|}{R}             & F1            \\ \hline
RNP*                     & 10 & \multicolumn{1}{c|}{91.5} & 40.0            & \multicolumn{1}{c|}{20.3}          & 25.2          & 10 & \multicolumn{1}{c|}{84}            & 49.1        & \multicolumn{1}{c|}{28.7}          & 32.0            & 10 & \multicolumn{1}{c|}{80.3}          & 38.6          & \multicolumn{1}{c|}{31.1}          & 29.7          \\
INVRAT*                  & 10 & \multicolumn{1}{c|}{91.0}   & 56.4          & \multicolumn{1}{c|}{27.3}          & 36.7          & 10 & \multicolumn{1}{c|}{90.0}            & 49.6        & \multicolumn{1}{c|}{27.5}          & 33.2          & 10 & \multicolumn{1}{c|}{80.0}            & 42.2          & \multicolumn{1}{c|}{32.2}          & 31.9          \\
A2R*                     & 10 & \multicolumn{1}{c|}{91.5} & 55.0          & \multicolumn{1}{c|}{25.8}          & 34.3          & 10 & \multicolumn{1}{c|}{85.5}          & 61.3        & \multicolumn{1}{c|}{34.8}          & 41.2          & 10 & \multicolumn{1}{c|}{80.5}          & 40.1          & \multicolumn{1}{c|}{34.2}          & 34.6          \\
FR*                      & 10 & \multicolumn{1}{c|}{93.5} & 51.9          & \multicolumn{1}{c|}{25.1}          & 31.8          & 10 & \multicolumn{1}{c|}{88.0}            & 54.8        & \multicolumn{1}{c|}{33.7}          & 39.5          & 10 & \multicolumn{1}{c|}{82.0}            & 44.3          & \multicolumn{1}{c|}{32.5}          & 33.7          \\
CR*                      & 10 & \multicolumn{1}{c|}{92.4} & 59.7          & \multicolumn{1}{c|}{31.6}          & 39.0          & 10 & \multicolumn{1}{c|}{86.5}          & 68.0        & \multicolumn{1}{c|}{42.0}          & 49.1          & 10 & \multicolumn{1}{c|}{82.5}          & 44.7          & \multicolumn{1}{c|}{37.3}          & 38.1          \\ \hline
PLMR$_{-L_{match}}$               & 10 & \multicolumn{1}{c|}{92.3} & 70.0          & \multicolumn{1}{c|}{45.3}          & 55.0          & 10 & \multicolumn{1}{c|}{86.3}          & 76.6        & \multicolumn{1}{c|}{48.6}          & 59.5          & 10 & \multicolumn{1}{c|}{80.2}          & 44.1          & \multicolumn{1}{c|}{50.5}          & 47.1          \\
PLMR                     & 10 & \multicolumn{1}{c|}{\textbf{98.2}} & \textbf{74.2} & \multicolumn{1}{c|}{\textbf{46.0}} & \textbf{56.8} & 10 & \multicolumn{1}{c|}{85.5} & \textbf{78.4} & \multicolumn{1}{c|}{\textbf{54.1}} & \textbf{64.0} & 10 & \multicolumn{1}{c|}{\textbf{85.4}} & \textbf{49.0} & \multicolumn{1}{c|}{\textbf{51.5}} & \textbf{50.2} \\ \hline
\end{tabularx}
\end{table*}

\section{EXPERIMENTS}
\subsection{Datasets}
    We use two widely used datasets in rationalization \cite{dmr,liu2022fr,dr,hu2024learning}. \textbf{BeerAdvocate} \cite{mcauley2012learning} is a dataset for multi-aspect sentiment prediction on beer reviews, where users rate each aspect on a scale of $[0,1]$. Following previous work \cite{invrat,inter-rat,liu2024d}, we use the original dataset with highly correlated aspects (referred to as the correlated dataset) to validate the effectiveness of PLMR in addressing severe rationalization degeneration and failure. \textbf{HotelReview} \cite{wang2010latent} is another dataset for multi-aspect sentiment prediction on hotel reviews, where each aspect is also rated on a scale of $[0,1]$. 

Consistent with prior work, we binarize the labels in datasets, with ratings $\geq 0.6$ as positive and ratings $\leq 0.4$ as negative.
Both datasets contain human-annotated rationales as the test set. 

\subsection{Baselines, Implementations, and Metrics}
\subsubsection{Baselines.}
To validate the application of our method in a rationalization framework for PLMs, we compared it with four baselines: rationalizing neural prediction (RNP \cite{lei2016rationalizing}), invariant rationalization (INVART \cite{invrat}), folded rationalization (FR \cite{liu2022fr}), and causal rationalization (CR \cite{cr}), which also utilize Bert-base-uncased. 
In the RNP, INVART, and FR frameworks, GRU was originally used. Here, we re-experiment using BERT.
In addition, to demonstrate the great improvement in the interpretation performance of our method, we compare it with the latest methods that can provide the best rationale using GRU, such as folded rationalization (FR \cite{liu2022fr}), Minimum Conditional Dependence (MCD \cite{liu2024d}), and Guidance-based Rationalization method (G-RAT \cite{hu2024learning}).

\subsubsection{Implementations.}
For a fair comparison, we adopted the same settings as previous work.
The specific implementation details are provided in the appendix \ref{Implementation-detail}.
In both the Dim-reduction layers and the predictor, the MLP employs a single linear layer, which is consistent with previous methods.
In PLMR, layers $0-l$ of PLM are the generator, which contains $m$ layers of Dim-reduction layers. In the main experiment, we validate the effectiveness of the method using Bert-base-uncased (12-layer transformers), where $l = 7$ and $m = 2$.
Additionally, we analyze the effect of different $l$ and $m$ values on the quality of the rationale through several experiments.
We also use other PLMs such as ELECTRA-base \cite{clark2020electra}, Roberta-base \cite{Liu2019RoBERTaAR}, and Bert-large, ELECTRA-large, Roberta-large with more extensive parameters (24-layer transformers) for supplementary experiments.
To demonstrate the effectiveness of the rationale selection module and rationale prediction module, we remove $L_{\text {match }}$ in the prediction module to serve as an ablation experiment.

\subsubsection{Metrics.}
As with previous methods, we focus on the quality of selected rationales. The test set includes human-annotated tokens, and we measure the alignment between the selected rationales and the human-annotated tokens using token-level precision (P), recall (R), and F1 score. Acc represents the precision of the prediction task using the selected rationales. S denotes the average proportion of selected tokens to the original text.

\subsection{Results}
\subsubsection{Comparison with state-of-the-art.}
We compare PLMR with rationalization frameworks using Bert and using GRU, respectively.

Table \ref{tab:bert-result} shows the results with previous methods using BERT. Under the same experimental settings, our method achieved significant improvements in F1 scores across all three aspects of beer reviews, with an increase of up to 31\% in the appearance aspect. 
This huge improvement shows the superiority of PLMR in solving severe rationalization degeneration and failure, which enables rationalization to be applied in PLMs and achieve ideal performance. Appendix \ref{rationales} visualizes some rationales from RNP and PLMR.

Table \ref{tab:gru-beer-result} shows the results with previous methods using GRU. We set the rationale sparsity $S$ to approximately 10\%, 20\%, and 30\% for comparison. As shown in Table \ref{tab:gru-beer-result}, we achieved significant improvements in F1 scores across various aspects.
Therefore, compared to these highly effective GRU-based models, our Bert-based method PLMR still provides better explanations, aligning with the expected outcomes of using PLMs. Table \ref{tab:gru-hotel-result} presents the experimental results on the HotelReview dataset using GRU. In this dataset, we set the rationale sparsity close to the human-annotated rationales.

These results demonstrate that our method not only addresses the degeneration and failure in rationalization within PLMs but also offers more accurate explanations than the existing methods.

\begin{table*}[htbp]
\caption{Comparison with GRU-Based Methods on BeerAdvocate Dataset. “ * ” represents the results from the paper of MCD \cite{liu2024d}. “ ** ” represents our reimplemented methods.}
\label{tab:gru-beer-result}
\begin{tabularx}{0.95\textwidth}{c|*{5}{>{\centering\arraybackslash}X}|*{5}{>{\centering\arraybackslash}X}|*{5}{>{\centering\arraybackslash}X}}
\hline
\multirow{2}{*}{Methods} & \multicolumn{5}{c|}{Appearance}                                                           & \multicolumn{5}{c|}{Aroma}                                                                & \multicolumn{5}{c}{Palate}                                                                \\ \cline{2-16} 
                         & S    & \multicolumn{1}{c|}{ACC}           & P             & R             & F1            & S    & \multicolumn{1}{c|}{ACC}           & P             & R             & F1            & S    & \multicolumn{1}{c|}{ACC}           & P             & R             & F1            \\ \hline
RNP*  & 10.0 & \multicolumn{1}{c|}{--} & 32.4 & 18.6& 23.6   & 10.0   & \multicolumn{1}{c|}{--}  & 44.8  & 32.4 & 37.6& 10.0 & \multicolumn{1}{c|}{--}& 24.6    & 23.5 & 24.0  \\
FR* & 11.1 & \multicolumn{1}{c|}{75.8}& 70.4& 42.0 & 52.6& 9.7  & \multicolumn{1}{c|}{87.7}& 68.1 & 42.2 & 52.1& 11.7 & \multicolumn{1}{c|}{87.9} & 43.7& 40.9& 42.3 \\
G-RAT**&10.5 & \multicolumn{1}{c|}{82.4}  &81.8 &46.3&59.1 &10.5& \multicolumn{1}{c|}{85.2} &82.0&55.4&66.2&9.5& \multicolumn{1}{c|}{89.2} &56.2 & 43.1& 48.8 \\
MCD* & 9.5  & \multicolumn{1}{c|}{81.5}& 94.2 & 48.4& 63.9& 9.9  & \multicolumn{1}{c|}{87.5} & 84.6 & 53.9& 65.8& 9.4  & \multicolumn{1}{c|}{87.3}& 60.9 & 47.1& 53.1\\
PLMR & 10.4 & \multicolumn{1}{c|}{\textbf{85.6}} & \textbf{96.5} & \textbf{54.9} & \textbf{70.0} & 10.6 & \multicolumn{1}{c|}{\textbf{92.9}} & \textbf{86.0} & \textbf{57.8} & \textbf{69.2} & 10.4 & \multicolumn{1}{c|}{\textbf{92.6}} & \textbf{62.8} & \textbf{51.5} & \textbf{56.6} \\ \hline
RNP*                      & 20.0   & \multicolumn{1}{c|}{--}            & 39.4          & 44.9          & 42.0            & 20.0   & \multicolumn{1}{c|}{--}            & 37.5          & 51.9          & 43.5          & 20.0   & \multicolumn{1}{c|}{--}            & 21.6          & 38.9          & 27.8          \\
FR*                       & 20.9 & \multicolumn{1}{c|}{84.6}          & 74.9          & 84.9          & 79.6          & 19.5 & \multicolumn{1}{c|}{89.3}          & 58.7          & 73.3          & 65.2          & 20.2 & \multicolumn{1}{c|}{88.2}          & 36.6          & 59.4          & 45.3          \\
G-RAT**&19.7& \multicolumn{1}{c|}{85.0} &80.2&85.2&82.6&20.2 & \multicolumn{1}{c|}{88.1}&60.5&78.2&68.2&20.3 & \multicolumn{1}{c|}{86.1} & 38.4 &62.7&47.6 \\
MCD*  & 20.0   & \multicolumn{1}{c|}{85.5}          & 79.3          & 85.5          & 82.3          & 19.3 & \multicolumn{1}{c|}{88.4}          & 65.8          & 81.4          & 72.8          & 19.6 & \multicolumn{1}{c|}{87.7}          & 41.3          & 65.0          & 50.5          \\
PLMR          & 19.5 & \multicolumn{1}{c|}{\textbf{86.3}} & \textbf{86.2} & \textbf{92.0} & \textbf{89.0} & 19.5 & \multicolumn{1}{c|}{\textbf{89.5}} & \textbf{74.2} & \textbf{91.4} & \textbf{81.9} & 19.7 & \multicolumn{1}{c|}{\textbf{90.4}} & \textbf{45.8} & \textbf{71.4} & \textbf{55.8} \\ \hline
RNP* & 30.0   & \multicolumn{1}{c|}{--} & 24.2          & 41.2  & 30.5     & 30.0   & \multicolumn{1}{c|}{--}            & 27.1          & 55.7          & 36.4          & 30.0   & \multicolumn{1}{c|}{--}            & 15.4          & 42.2          & 22.6          \\
FR*                       & 29.6 & \multicolumn{1}{c|}{86.4}          & 50.6          & 81.4          & 62.3          & 30.8 & \multicolumn{1}{c|}{88.1}          & 37.4          & 75.0            & 49.9          & 30.1 & \multicolumn{1}{c|}{87.0}            & 24.5          & 58.8          & 34.6          \\
G-RAT**&29.6  & \multicolumn{1}{c|}{87.2}& 56.0 &89.4 &68.9 &29.8 & \multicolumn{1}{c|}{90.4} &42.4&81.1&55.7& 29.7& \multicolumn{1}{c|}{86.2}  &27.0 &64.4 &38.0     \\
MCD*                      & 29.7 & \multicolumn{1}{c|}{86.7}          & 59.6          & 95.6          & 73.4          & 29.6 & \multicolumn{1}{c|}{90.2}          & 46.1          & 87.5          & 60.4          & 29.4 & \multicolumn{1}{c|}{87.0}            & 30.5          & 72.4          & 42.9          \\
PLMR                     & 29.3 & \multicolumn{1}{c|}{\textbf{87.5}} & 59.3          & 94.8          & 72.9          & 29.2 & \multicolumn{1}{c|}{\textbf{93.0}} & \textbf{50.6} & \textbf{92.9} & \textbf{65.5} & 29.5 & \multicolumn{1}{c|}{\textbf{90.1}} & \textbf{34.4} & \textbf{80.1} & \textbf{48.1} \\ \hline
\end{tabularx}
\end{table*}

\begin{table}[htbp]
\caption{Results on HotelReview. “ * ” represents the results from MCD \cite{liu2024d}. “ ** ” represents our re-implementations.}
\label{tab:gru-hotel-result}
\begin{tabularx}{0.4\textwidth}{c|*{2}{>{\centering\arraybackslash}X}|*{3}{>{\centering\arraybackslash}X}}
\hline
Method  & S     & ACC  & P     & R    & F1   \\ \hline
RNP*    & 11.0  & 97.5 & 34.2  & 32.9 & 33.5 \\
FR*     & 11.5  & 94.5 & 44.8  & 44.7 & 44.8 \\
G-RAT** & 12.1  & 97.9 & 44.4  & 47.1 & 46.3 \\
MCD*    & 11.8  & 97.0 & 47.0  & 48.6 & 47.8 \\
PLMR    & 12.8  & \textbf{98.3} & 46.5  & \textbf{52.3} & \textbf{49.2} \\ \hline
\end{tabularx}
\end{table}

\subsubsection{Ablation study.}
To validate the effectiveness of the rationale selection module and rationale prediction module, we remove the $L_{x}$ and $L_{match}$ in the rationale prediction module during training. As the line PLMR$_{-L_{match}}$ in Table \ref{tab:bert-result}, excluding the regularization term $L_{match}$, $L_{x}$ decreases the F1 score, indicating that the effectiveness of $L_{match}$ and $L_{x}$ in the prediction module. 
However, the decreased F1 score is still significantly higher than that of other methods, which demonstrates that the rationale selection module can select relatively correct rationales, avoiding rationalization degeneration and failure.
Simultaneously, we note that the F1 score in appearance does not decrease significantly. That is because the golden rationale in appearance tends to appear early in the text and is less impacted by spurious correlation. As a result, there has been a slight improvement in the performance of $L_{\text {match }}$.

\subsubsection{Analysis of the Generator's Layers $l$.} 
\begin{figure}[htbp]
    \centering
    \begin{subfigure}{0.23\textwidth}
        \centering
        \includegraphics[width=\textwidth]{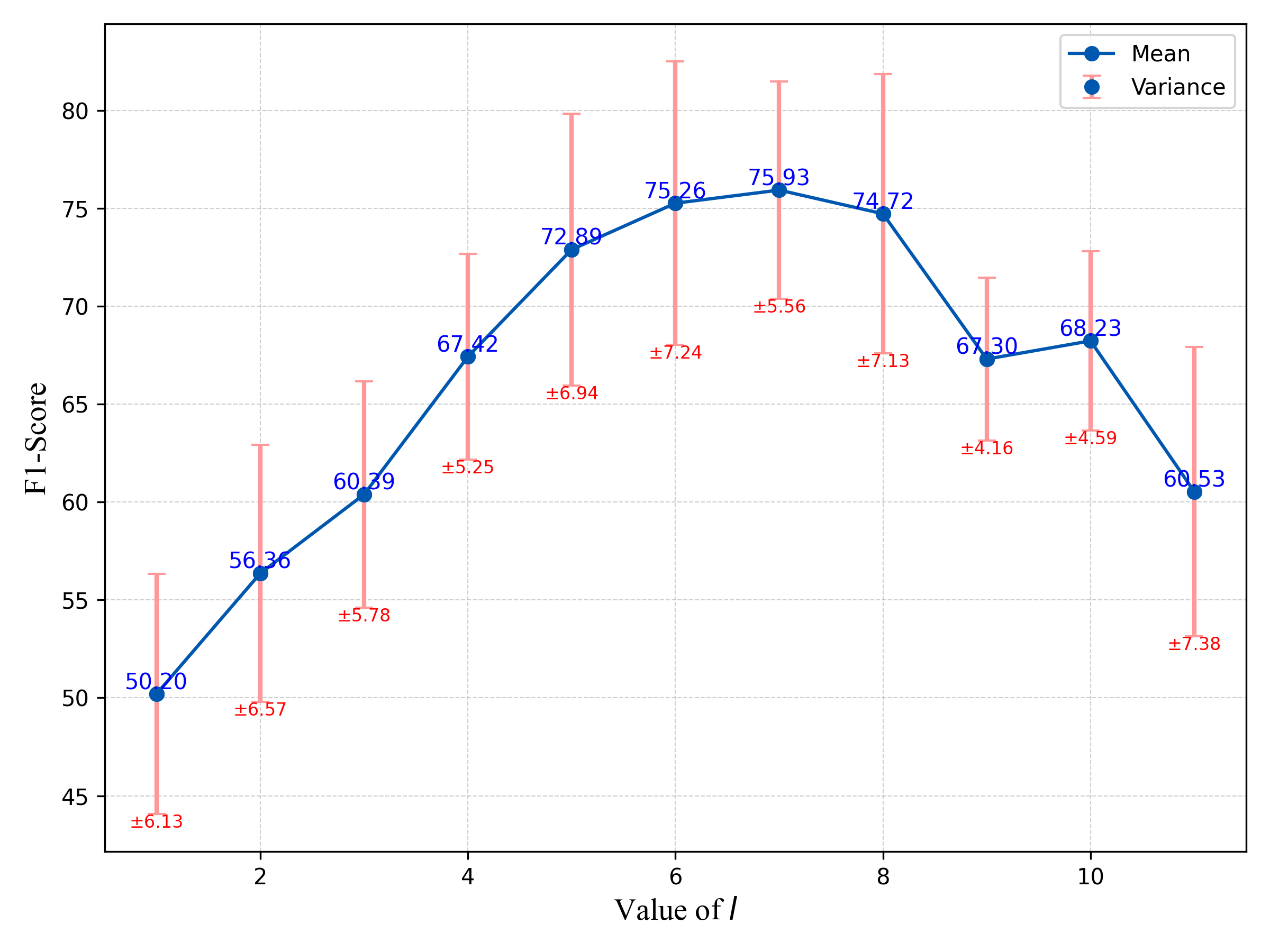}
        \caption{BERT}
        \label{figure-diff-l1}
    \end{subfigure}
    \hfill
    \begin{subfigure}{0.23\textwidth}
        \centering
        \includegraphics[width=\textwidth]{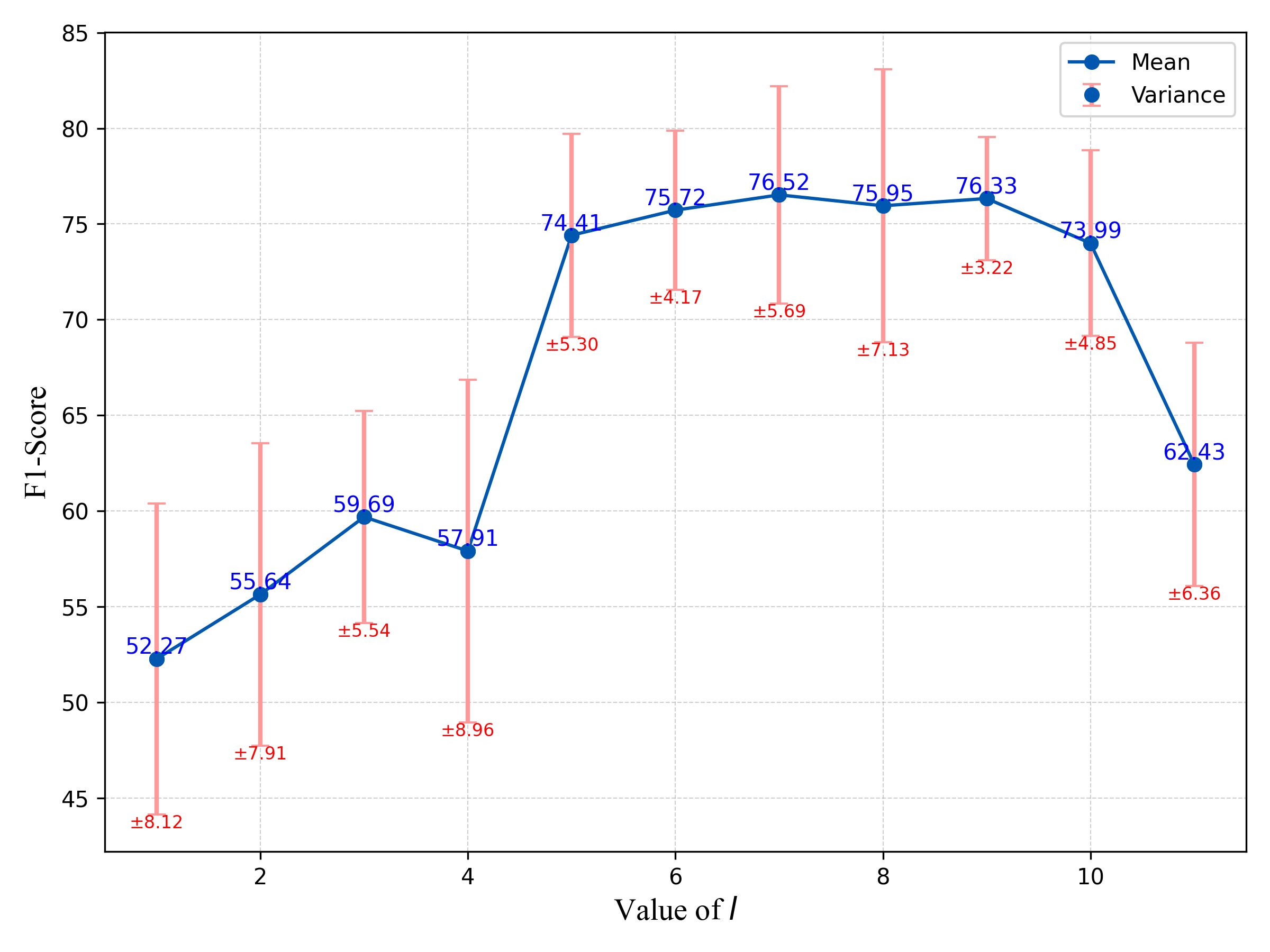}
        \caption{ELECTRA}
        \label{figure-diff-l2}
    \end{subfigure}
    \caption{Analysis of the generator's layers $l$. The F1 score is averaged over five different random seeds. The rationale selection sparsity is approximately 0.2.}
    \label{figure-diff-l}
\end{figure}

How should the number of layers in the generator and predictor be allocated to optimize the selected rationale?
To analyze the effect of the generator's layers $l$, we keep the number of Dim-reduction layers $m = 1$ unchanged, and the generator's layer $l$ from 1 to 11, with the number of layers in the predictor corresponding to $12 - l$. Experiments were conducted on the aroma aspect of the beer review dataset using BERT-base and ELECTRA-base, respectively.
As shown in Figure \ref{figure-diff-l}, both smaller and larger $l$ values will result in a decrease in the F1 score.
The reason is that the generator and predictor work together to extract rationales and make task predictions in selective rationalization.
A small $l$ value can result in an excessive number of layers in the predictor, leading to overfitting on incorrect rationales. Conversely, a large $l$ value increases the homogeneity among tokens, which similarly reduces the quality of the selected rationale.
Therefore, in Figure \ref{figure-diff-l1}, the F1 score in PLMR reaches the maximum when $l=7$. 
At this point, the layers of the generator and the predictor achieve a balance, allowing the selection of the optimal rationale.
The results in Figure \ref{figure-diff-l2} validate the above analysis as well.

\subsubsection{Analysis of the Dim-reduction layers.}
\begin{figure}[htbp]
    \centering
    \begin{subfigure}{0.23\textwidth}
        \centering
        \includegraphics[width=\textwidth]{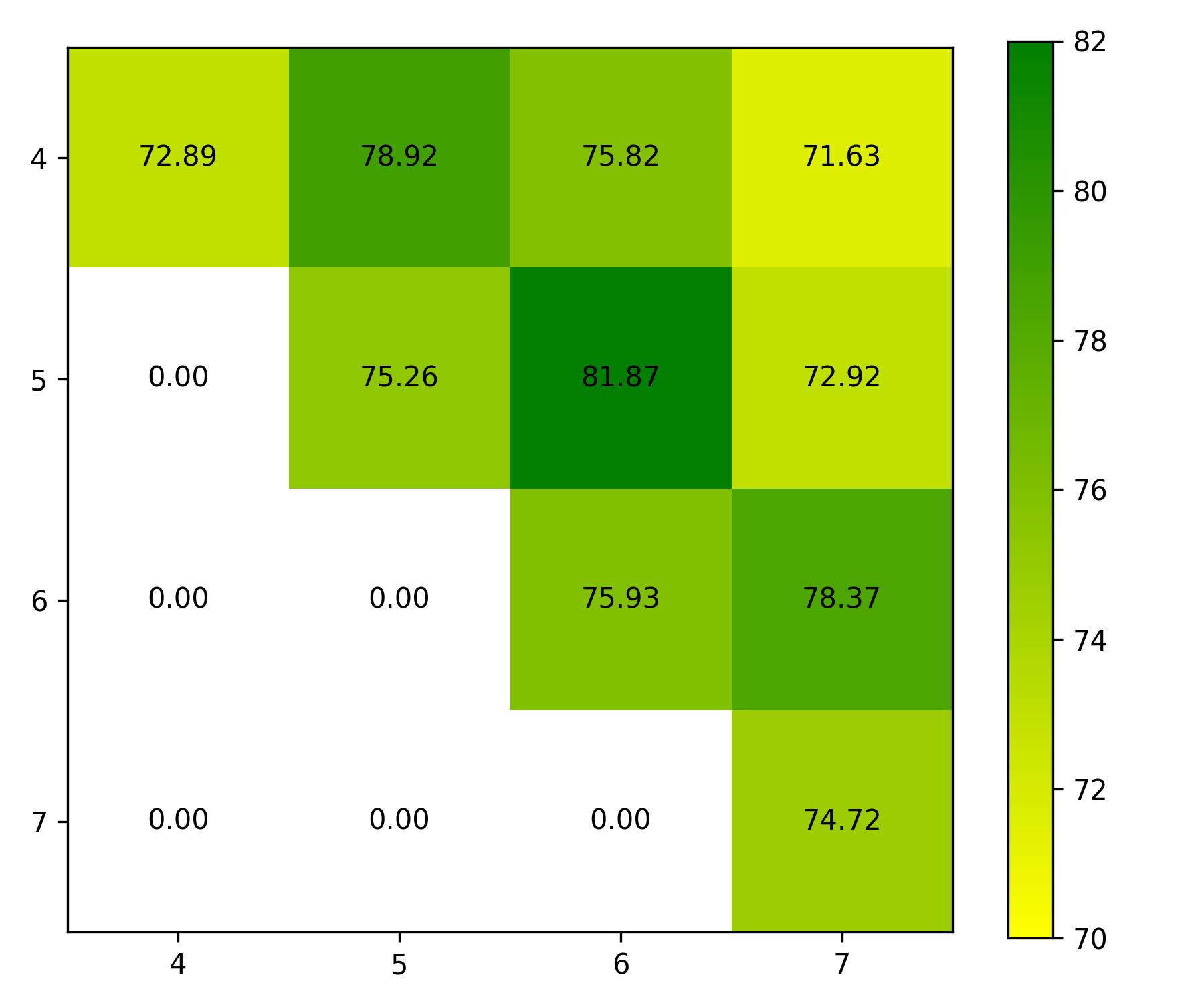}
        \caption{BERT-base}
        \label{figure-diff-m1}
    \end{subfigure}
    \hfill
    \begin{subfigure}{0.23\textwidth}
        \centering
        \includegraphics[width=\textwidth]{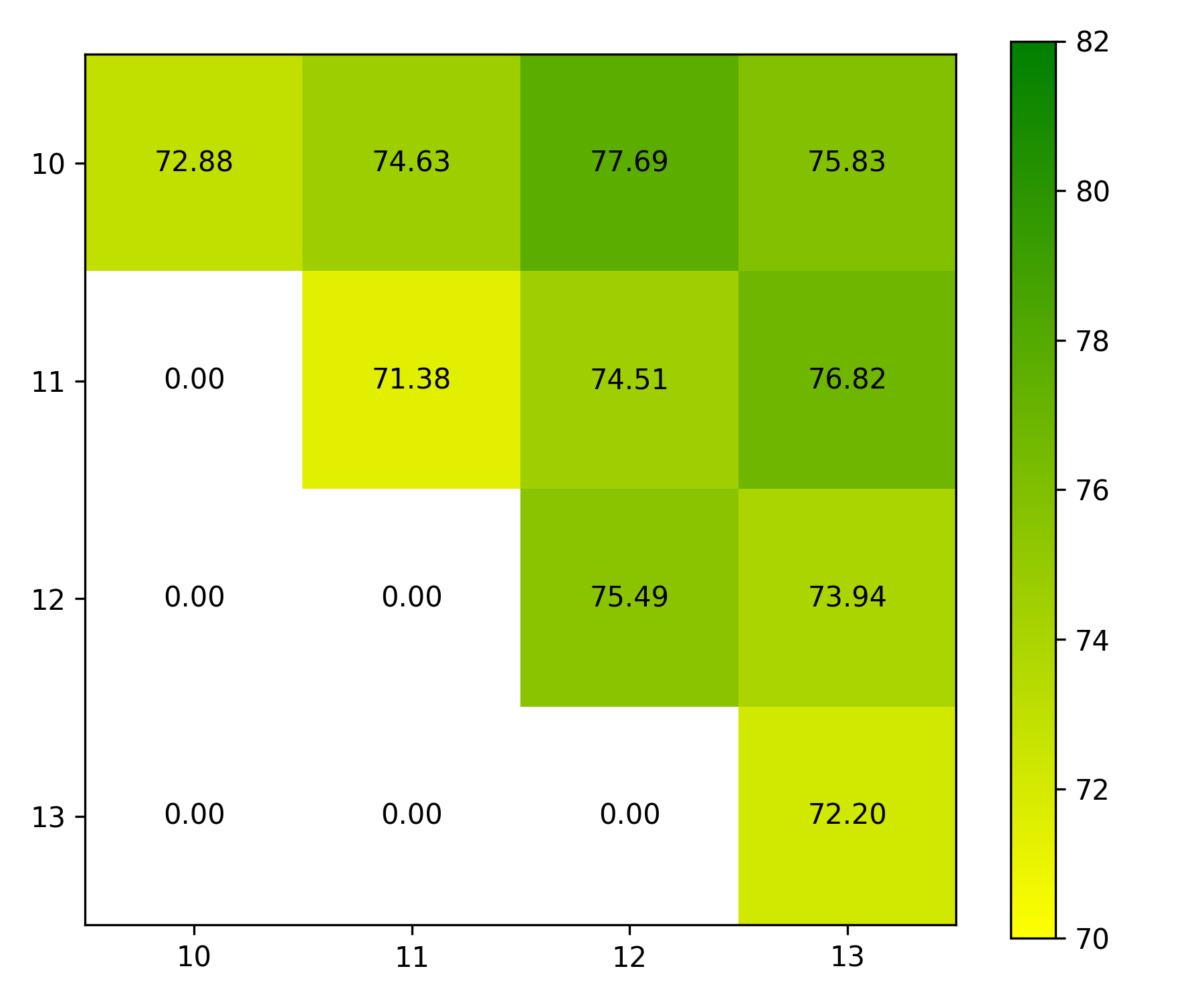}
        \caption{BERT-large}
        \label{figure-diff-m2}
    \end{subfigure}
    \caption{Analysis of different Dim-reduction layers.}
    \label{figure-diff-m}
\end{figure}

Based on the above analysis, selecting rationales in the middle layers yields better results. Therefore, we analyze the impact of Dim-reduction layers in the middle four layers of BERT-base and BERT-large, respectively. 
As shown in Figure \ref{figure-diff-m}, the value at coordinates (x, y) represents the F1 score for the Dim-reduction layers from layer y to layer x. Figure \ref{figure-diff-m1} shows that using two Dim-reduction layers in BERT-base results in a much higher F1 score compared to using just one layer. This demonstrates the effectiveness of multi-layer context pruning.
Additionally, it is observed that a greater number of Dim-reduction layers (3 layers) are required with BERT-large to reach optimal rationale quality (F1-score) in Figure \ref{figure-diff-m2}. 
This is because Bert-large has more layers, and a larger generator will make rationale selection more difficult, thus requiring more Dim-reduction layers.
This observation indirectly confirms the utility of Dim-reduction layers in pruning irrelevant tokens.
This experiment can also be used as an ablation experiment to analyze the Dim-reduction layers.

\subsubsection{Analysis of different pre-trained language models.}
To validate the effectiveness of our method on various PLMs, we conduct extensive experiments on different models. As shown in Table \ref{tab:plms-result}, we additionally perform experiments on ELECTRA-base and Roberta-base using the same $l = 7$ and $m = 2$. 
We also conduct experiments with larger models, such as BERT-large, ELECTRA-large, and Roberta-large. Based on the previous analysis, more Dim-reduction layers are required to generate the optimal rationale in these models. Therefore, we set $m = 3$ and $l = 13$.
In Table \ref{tab:plms-result}, the larger-parameter PLMs still provide highly effective rationales when using the PLMR framework.
The results demonstrate that our method effectively applies rationale selection to PLMs, enabling them to provide faithful and human-understandable rationales while completing NLP tasks.
\begin{table}[htbp]
\caption{The result of different PLMs. The dataset is the Appearance aspect of the BeerAdvocate dataset.}
\label{tab:plms-result}
\begin{tabularx}{0.4\textwidth}{c|*{2}{>{\centering\arraybackslash}X}|*{3}{>{\centering\arraybackslash}X}}
\hline
PLM    & S & ACC & P & R & F1 \\ \hline
BERT-base    &19.5   &86.3     &86.2   &92.0   &89.0    \\
ELECTRA-base &19.4   &87.6     &86.5   &91.8   &89.1    \\
Roberta-base &19.6  &84.2  &84.6 &90.7 &87.5 \\
BERT-large    &19.3 &92.1 &85.9 &90.7 &88.3 \\
ELECTRA-large &19.4   &90.9    &86.7   &92.1   &89.3    \\
Roberta-large &19.7  &91.7 &86.0 &92.3 &89.0 \\\hline
\end{tabularx}
\end{table}

\subsubsection{Parameters sensitivity analysis.}
In the previous setup, we set $\lambda = 1.0$ in $L_{\text {match }}$. To gain an insight into the effects of selecting different $\lambda$. We further conduct experiments varying $\lambda$ from 0.1 to 10. The details are in Appendix \ref{para-analysis}.

\section{Conclusion and future work}
We propose the method PLMR to address the challenge of applying selective rationalization in pre-trained language models. Specifically, PLMR reduces the homogeneity among tokens in the generator, making it easier for the generator to select meaningful tokens. Simultaneously, it uses full-text information to regularize the predictor.
Experiments on multiple pre-trained language models have demonstrated the superior performance of PLMR.
In addition, several ablation experiments have validated the effectiveness of the generator, the predictor, and the Dim-reduction layers.
This method enables the initial application of selective rationalization in PLMs to achieve optimal performance. 
Although our method, like previous studies, focuses on classification problems, as an interpretable framework, our work can also be applied to other types of tasks, such as regression and unsupervised clustering.
We can also explore other methods to address degeneration and failure further to advance rationalization research in PLMs.
Additionally, investigating whether PLMR can provide robustness to adversarial attacks by selecting precise rationales is also significant research in future.
Our current work is an essential step toward building interpretable methods scalable to more complex models, and we plan to explore LLM explainability in future work.

\begin{acks}
This work is supported by the National Science and Technology Major Project of China (2021ZD0111801) and the National Natural Science Foundation of China (under grant 62376087).
\end{acks}

\bibliographystyle{ACM-Reference-Format}
\balance
\bibliography{reference}

\appendix

\section{Detailed Methods Analysis}

\begin{table*}[]
\caption{Examples of generated rationales from BERT-RNP and PLMR. Human-annotated rationales are \ul{underlined}. Rationales selected by the generator are highlighted in \textcolor{red}{red}.}
\label{tab:bert-rnp-plmr}
\centering
\begin{tabular}{p{8cm}p{8cm}}
\hline
BERT-RNP & PLMR(ours) \\ \hhline{:=::=:}
\textbf{Aspect:} Beer-Aroma &\textbf{Aspect:} Beer-Aroma  \\
\textbf{Label:} Positive, \textbf{Pred:} Positive & \textbf{Label:} Positive, \textbf{Pred:} Positive \\
\textbf{Text:} this \textcolor{red}{beer} poured out as \textcolor{red}{a} nice golden color with a white head on top . the head retention was pretty good on this brew . \ul{i }\ul{found the smell of the beer to have a nice aroma }\textcolor{red}{\ul{of }}\ul{caramel and some }\textcolor{red}{\ul{light hops}}\ul{ on the nose .} \textcolor{red}{the} taste of the beer was really nicely done \textcolor{red}{i} thought . the flavors of sweet malts and \textcolor{red}{the} bitter finish \textcolor{red}{worked well} together . the mouthfeel \textcolor{red}{was} very drinkable and \textcolor{red}{could} easlier have serval back to \textcolor{red}{back . overall} this \textcolor{red}{brew} is pretty good \textcolor{red}{and} i would n't nmind drinking \textcolor{red}{it} again one day .&
\textbf{Text:} this beer poured out as a nice golden color with a white head on top . the head retention was pretty good on this brew . \ul{i found }\textcolor{red}{\ul{the smell of the beer to have a nice aroma of caramel and some light hops on the nose .} the} taste of the beer was really nicely done i thought . the flavors of sweet malts and the bitter finish worked well together . the mouthfeel was very drinkable and could easlier have serval back to back . overall this brew is pretty good and i would n't nmind drinking it again one day .\\
\hline
\end{tabular}
\end{table*}

\subsection{Homogeneity of tokens}
\label{homogeneity}
\begin{figure}[htbp]
    \centering
    \begin{subfigure}{0.23\textwidth}
        \centering
        \includegraphics[width=\textwidth]{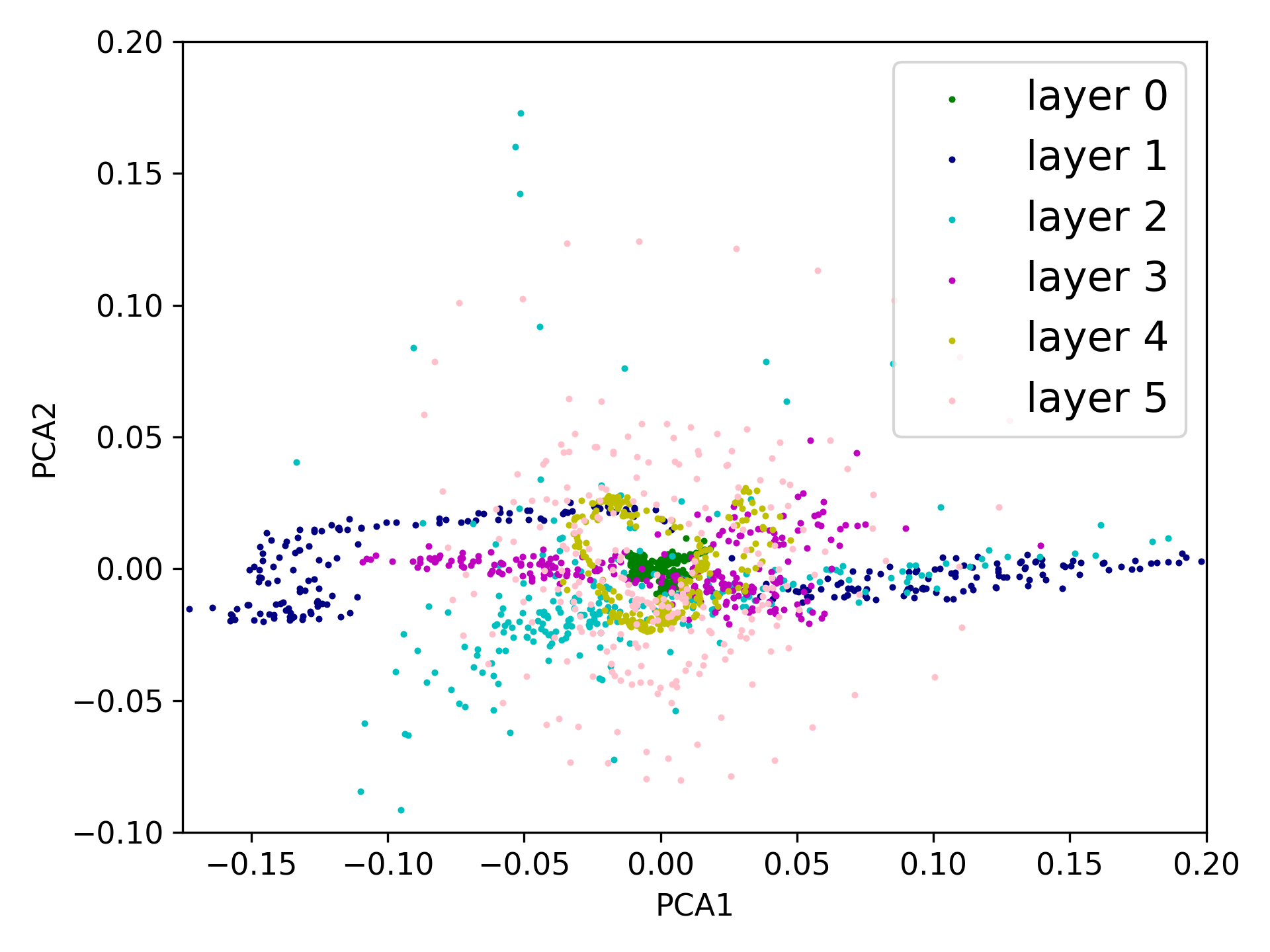}
        \caption{layers 0-5}
        \label{layer0-5}
    \end{subfigure}
    \hfill
    \begin{subfigure}{0.23\textwidth}
        \centering
        \includegraphics[width=\textwidth]{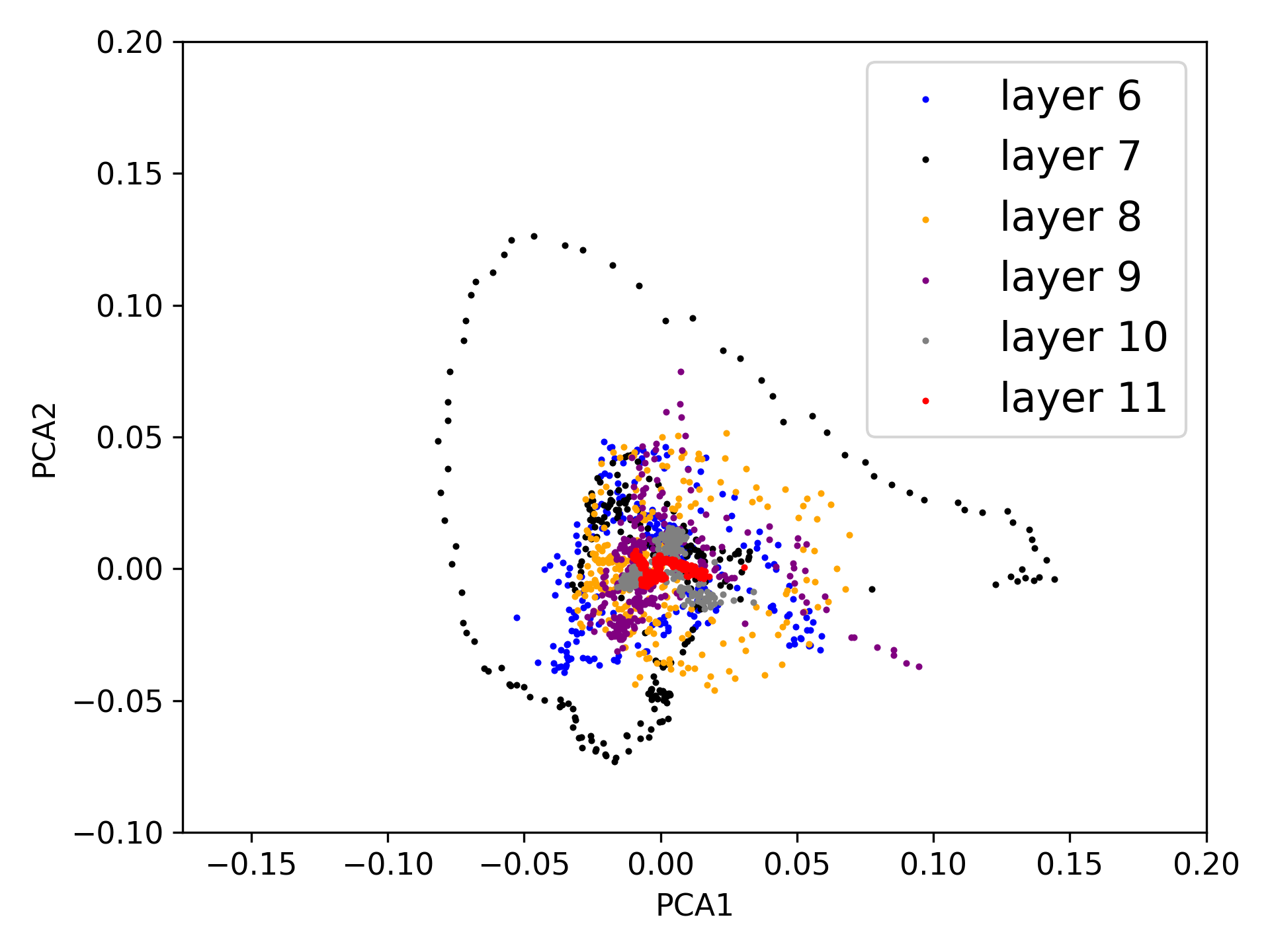}
        \caption{layers 6-11}
        \label{layer6-11}
    \end{subfigure}
    \caption{Visualization of attention-weight vectors in the first attention head. Dimensionality reduction using PCA.
    (a): Distribution of attention-weight vectors of tokens from layers 0-5.
    (b): Distribution of attention-weight vectors of tokens from layers 6-11.
    }
    \label{figure-pca}
\end{figure}

The Transformer employs a multi-head self-attention mechanism to enable the model to obtain information from different representation subspaces. The $h$-th self-attention head is described as follows:
\begin{equation}
  Q_{h}=X W_{h}^{Q}, K=X W_{h}^{K}, V=X W_{h}^{V}.
\end{equation}
\begin{equation}
  A_{h}=\operatorname{soft} \max \left(\frac{Q_{h} K_{h}^{T}}{\sqrt{d_{k}}}\right).
  \label{eq:Ah}
\end{equation}
\begin{equation}
  H_{h}=\operatorname{AttentionHead}(X)=A_{h} V_{h}.
  \label{eq:Hh}
\end{equation}
Therefore, we use the score $A_{i,j}^{h}$ to represent how much attention that token $x_{i}$ pays to token $x_{j}$ in the $h$-th attention head of the transformer.
The attention weight of token $x_{i}$ on all tokens in the $h$-th attention header can be expressed as the attention-weight vector $A_{i}^{h}=\left[A_{i,1}^{h}, \cdots, A_{i,n}^{h}\right]$.
Figures \ref{layer0-5} and \ref{layer6-11} illustrate the distribution of attention-weight vectors for different tokens in the first six layers and the last six layers of BERT, respectively.
We observed that in the initial layers, there are significant discrepancies in the distribution of the weight vectors.
In the final layers, the attention-weight vectors converge, reaching full clustering in the last layer with minimal distribution discrepancies.
The clustering of the final layer attention-weight vectors indicates that all tokens exhibit similar attention dependencies. This will result in a highly similar token representation, which we refer to as token homogeneity.


\subsection{Analysis of $L_{\text {match }}$}
\label{proof-lmatch}
First, we assume that $\lambda>0$ in Eq. \ref{l_match}.
In the early training phase, the generator selects poor rationales, yet the complete text encompasses all necessary information for better prediction support.
Thus we can obtain: $H(Y \mid R) \geq H(Y \mid X)$.
The entropy of the actual distribution influences the value of the cross-entropy loss; higher entropy in the actual distribution typically results in a higher cross-entropy loss for a given prediction distribution. 
So there is:
\begin{equation}
  H(Y \mid R) \geq H(Y \mid X) \Leftrightarrow L_{\text {task }}^{r} \geq L_{\text {task }}^{x}.
  \label{a}
\end{equation}
At this time, $x=L_{\text {task }}^{r}-L_{\text {task }}^{x} \geq 0$, and we expect to optimize the rationale quality by minimizing $L_{\text {match }}=\lambda h\left(x\right)$. Thus, with the guarantee that $\lambda>0$, $h(x)$ monotonically increases when $x \geq 0$.

In the later training phase, the generator selects superior rationales, while the complete text $X$ includes tokens that are irrelevant to the labels. So $H(Y \mid R) \leq H(Y \mid X)$. Similar to Equation \ref{a}:
\begin{equation}
  H(Y \mid R) \leq H(Y \mid X) \Leftrightarrow L_{\text {task }}^{r} \leq L_{\text {task }}^{x}.
\end{equation}
At this point $x=L_{\text {task }}^{r}-L_{\text {task }}^{x} \leq 0$. To ensure a smaller $L_{\text {task }}^{r}$, we also need the function $h(x)$ to be monotonically decreasing when $x \leq 0$.

So the function $h(x)$ needs to be monotonically increasing, e.g., $h(x)=x$ and $f(x)=a^{x} \quad(a>1)$.

\section{more results}
\subsection{The example in Figure \ref{figure-Bert-GRU}}
\label{examples1}

\begin{table}[]
\caption{Comparison of BERT and GRU Classifiers on Untrained Aspects.}
\label{figure3_example}
\centering
\begin{tabular}{p{8cm}p{8cm}}
\hline
BeerAdvocate \\ \hhline{:=:}
Train on the appearance aspect\\
\textbf{Label:} Appearance(0.6, positive), Aroma(0.4, negative), Palate(0.3, negative)\\
\textbf{Text:} a : poured a hazy amber color with a nice white cap s : faded piney hops , caramel t : caramel and faded hops battle for dominance while the metallic and old earthy notes fight for second place . bad . m : medium to full body and not one i would recommend hiolding on your palate o : poor ipa , drainpoured after two sips\\
\textbf{Prediction(Bert):} Appearance(positive), Aroma(negative), Palate(negative)\\
\textbf{Prediction(GRU):} Appearance(positive), Aroma(positive), Palate(positive)\\
\hline
\end{tabular}
\end{table}

To more clearly illustrate how PLMs like BERT can strengthen the impact of spurious correlations, we add the following example text in Table \ref{figure3_example}.

We trained classifiers using BERT and GRU, respectively, based on appearance. Next, predictions were made on three aspects of the text. In Table \ref{figure3_example}, the BERT-based classifier still made correct predictions on the untrained aspects of aroma and palate, while the GRU-based classifier made incorrect predictions.

\subsection{Parameters Sensitivity Analysis}
\label{para-analysis}
We conducted experiments on three aspects of the BeerAdvocate and the HotelReview Datasets.
As shown in Figure \ref{analysis}, both small and large values of $\lambda$ lead to lower F1 scores. 
According to the analysis in Appendix \ref{proof-lmatch}, we expect $x=L_{\text {task }}^{r}-L_{\text {task }}^{x} \geq 0$ when the rationale is poor and $x \leq 0$ when the rationale is good; however, a large $\lambda$ leads to a rapid decrease in $x$, which results in $x \leq 0$ when the rationale is poor.
Similarly, a smaller $\lambda$ leads to a slow decrease in $x$, leading to $x \geq 0$ when the rationale is better. thus neither large nor small $\lambda$ meets our expectations of $L_{\text {match }}$, or even seriously affects the rationalization.

\begin{figure}[htbp]
    \centering
    \begin{subfigure}{0.23\textwidth}
        \centering
        \includegraphics[width=\textwidth]{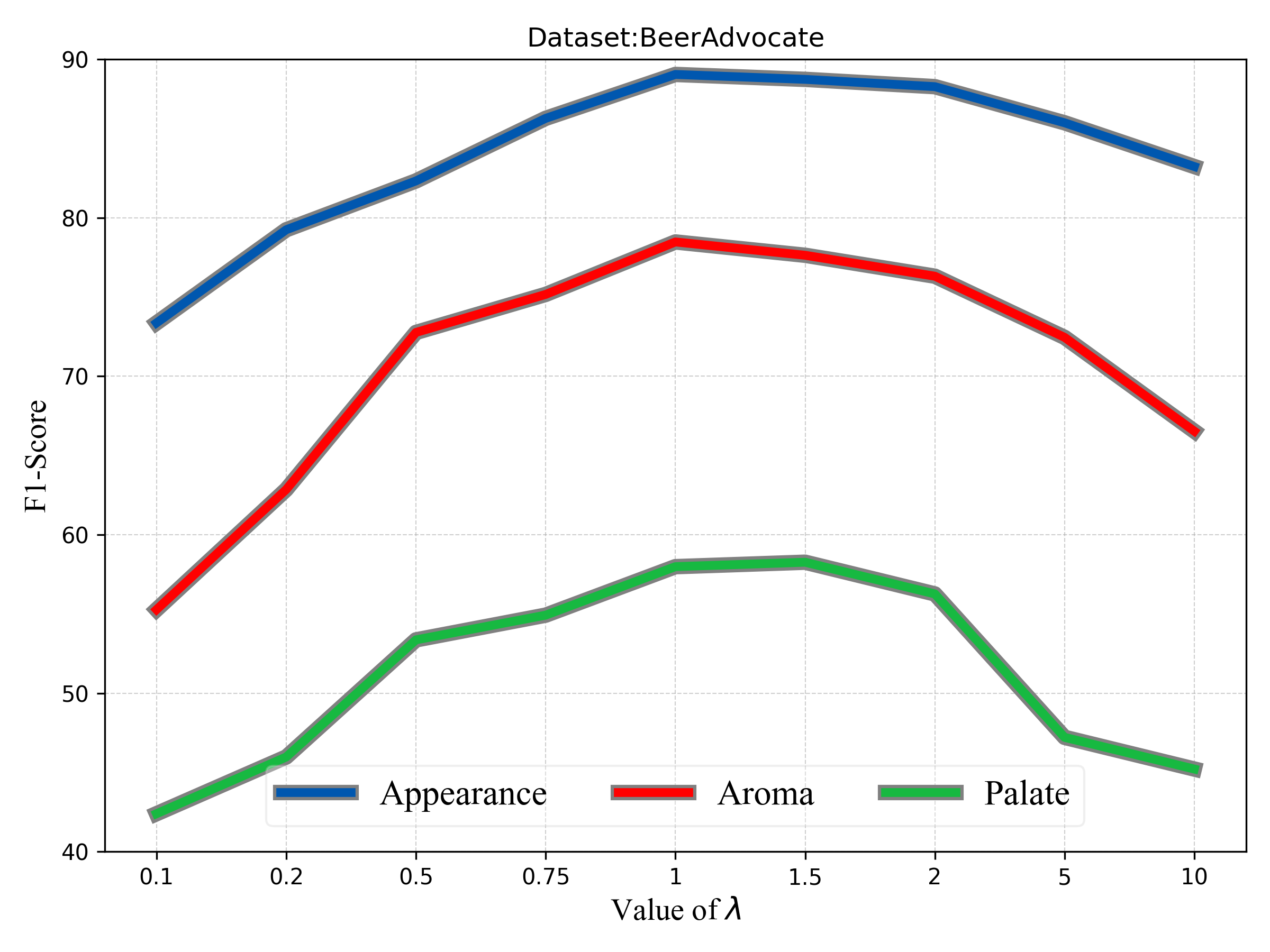}
        \caption{}
        \label{beer}
    \end{subfigure}
    \hfill
    \begin{subfigure}{0.23\textwidth}
        \centering
        \includegraphics[width=\textwidth]{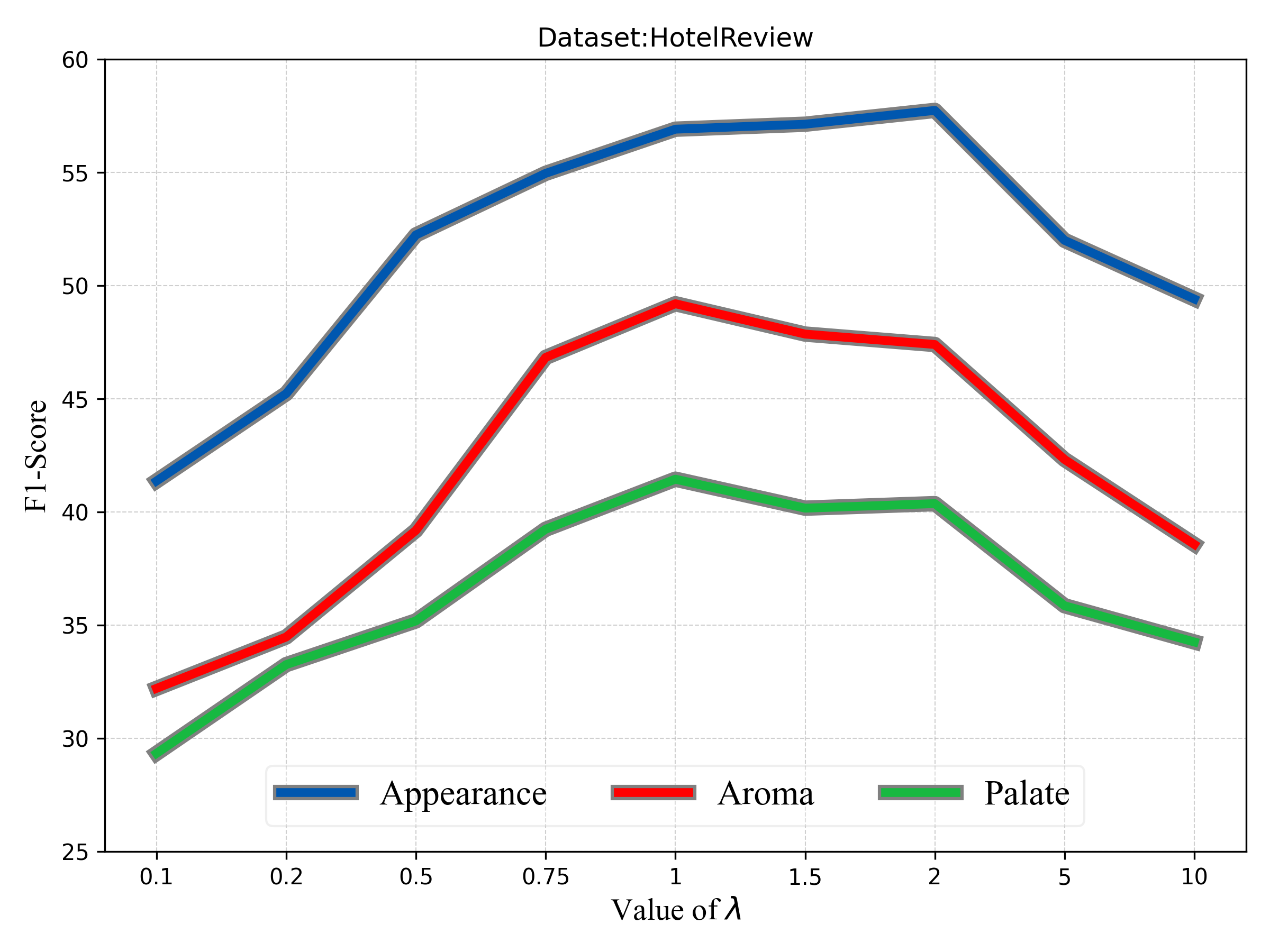}
        \caption{}
        \label{hotel}
    \end{subfigure}
    \caption{Analysis of the sensitivity of $\lambda$.
    }
    \label{analysis}
\end{figure}
\subsection{Rationale Selected by BERT-RNP and PLMR}
\label{rationales}
As shown in Table \ref{tab:bert-rnp-plmr}, we visualize the rationales selected by BERT-RNP and PLMR. BERT-RNP consistently selects meaningless tokens as rationales, whereas PLMR can choose relatively correct rationales.
More examples can be found in our source code.

\section{EXPERIMENTAL SETUP}
\subsection{Implementation Details}
\label{Implementation-detail}
During training, the learning rate was set to 5e-6 for transformer layers and 2e-5 for the MLP in the predictor and the dimension-reduction layer. The batch size was 64, and the maximum sequence length was 256. We trained for 30 epochs on the BeerAdvocate dataset and 100 epochs on the HotelReview dataset.
For hyperparameter tuning, parameter $\lambda$ was set to 1, and sparsity control $\alpha$ was adjusted within 0.1-0.3 to ensure rationale percentages of around 0.1, 0.2, or 0.3. $\lambda_{1}$ and $\lambda_{2}$ were also adjusted to find the optimal settings. 
For a fair comparison, the experiments in Table \ref{tab:bert-result} used a short-text version of the dataset by filtering the texts over a length of 120, as was done with CR.
All experiments were conducted using PyTorch on A100 GPUs.

\subsection{The setup of Table \ref{tab:bert-gru}}
\label{table1-detail}
The experimental results of BERT are derived from those of CR, which creates a short-text version of the dataset by filtering out texts exceeding a length of 120. For a fair comparison, we utilized the same filtered dataset for both the single-layer GRU and the three-layer GRU. We used a learning rate of 0.0001, a batch size of 128, and trained for 150 epochs on the Aroma aspect of the BeerAdvocate dataset.

\subsection{The setup of Table \ref{table-failure}}
\label{table2-detail}
In Table 2, we trained RNP models based on BERT and GRU, and then calculated the probability of rationale failure on the test set. The dataset used was the aroma aspect of BeerAdvocate, with a rationale selection sparsity of 0.2. The prediction accuracies for BERT-RNP and GRU-RNP were 0.82 and 0.78, respectively.
\end{CJK}
\end{document}